\documentclass{article}

\usepackage[main, final]{neurips_2026}
\usepackage[utf8]{inputenc} % allow utf-8 input
\usepackage[T1]{fontenc}    % use 8-bit T1 fonts
\usepackage{hyperref}       % hyperlinks
\usepackage{url}            % simple URL typesetting
\usepackage{booktabs}       % professional-quality tables
\usepackage{amsfonts}       % blackboard math symbols
\usepackage{nicefrac}       % compact symbols for 1/2, etc.
\usepackage{microtype}      % microtypography
\usepackage{xcolor}         % colors
\usepackage{amsmath}
\usepackage{hyperref}
\usepackage{algorithm}
\usepackage[most]{tcolorbox}
\usepackage{enumitem}
\usepackage{multirow}
\usepackage{xcolor}   
\usepackage[dvipsnames]{xcolor}
\usepackage{pifont}   
\usepackage{amsthm}
\usepackage{algpseudocode}  
\usepackage{wrapfig}
\usepackage{tabularx}
\usepackage[table]{xcolor}
\usepackage{fontawesome5}
\usepackage[dvipsnames]{xcolor}

\definecolor{cfblue7}{HTML}{3067FF}
\newcommand{\cfblue}[1]{{\color{cfblue7}#1}}

\usepackage[framemethod=TikZ]{mdframed}

\mdfdefinestyle{graybox}{
    backgroundcolor=gray!10,
    linecolor=gray!40,
    linewidth=0.5pt,
    roundcorner=4pt,
    innertopmargin=0.5em,
    innerbottommargin=0.5em,
    innerleftmargin=10pt,
    innerrightmargin=10pt,
    skipabove=0.4em,
    skipbelow=0.4em,
}

\usepackage{amsthm}
\usepackage{thmtools}
\hypersetup{
    colorlinks=true,
    linkcolor=RoyalBlue,
    citecolor=RoyalBlue,
}
\usepackage[capitalize,noabbrev]{cleveref}
\declaretheoremstyle[
    spaceabove=1em,
    spacebelow=1em,
    headfont=\normalfont\bfseries,    % "Remark N." → normal bold
    notefont=\normalfont,             % 괄호 안 제목 → normal (italic 아님)
    notebraces={(}{)},
    bodyfont=\itshape,                % 본문 → italic
    postheadspace=0.5em,
    headpunct={.},
]{spacedremark}

\declaretheoremstyle[
    spaceabove=1em,
    spacebelow=1em,
    headfont=\normalfont\bfseries,    % "Remark N." → normal bold
    notefont=\normalfont,             % 괄호 안 제목 → normal (italic 아님)
    notebraces={(}{)},
    bodyfont=\itshape,                % 본문 → italic
    postheadspace=0.5em,
    headpunct={.},
]{spacedproposition}
\declaretheoremstyle[
    spaceabove=1em,
    spacebelow=1em,
    headfont=\bfseries,
    notefont=\normalfont,
    notebraces={(}{)},
    bodyfont=\normalfont,
    postheadspace=0.5em,
    headpunct={.},
]{spaceddefinition}

\declaretheoremstyle[
    spaceabove=1em,
    spacebelow=1em,
    headfont=\bfseries,
    notefont=\normalfont,
    notebraces={(}{)},
    bodyfont=\itshape,
    postheadspace=0.5em,
    headpunct={.},
]{spacedplain}

\declaretheorem[name=Theorem,numberwithin=section,style=spacedplain,refname={Theorem,Theorems},Refname={Theorem,Theorems}]{theorem}
\declaretheorem[name=Proposition,sibling=theorem,style=spacedplain,refname={Proposition,Propositions},Refname={Proposition,Propositions}]{proposition}

\declaretheorem[name=Remark,sibling=theorem,style=spacedremark,refname={Remark,Remarks},Refname={Remark,Remarks}]{remark}
\crefname{appendix}{App.}{Apps.}
\Crefname{appendix}{App.}{Apps.}
\crefname{section}{Sec.}{Secs.}
\Crefname{section}{Sec.}{Secs.}
\crefname{equation}{Eq.}{Eqs.}
\Crefname{equation}{Eq.}{Eqs.}
\crefname{figure}{Fig.}{Figs.}
\Crefname{figure}{Fig.}{Figs.}
\crefname{table}{Tab.}{Tabs.}
\Crefname{table}{Tab.}{Tabs.}
\crefname{theorem}{Theorem}{Theorems}
\Crefname{theorem}{Theorem}{Theorems}
\crefname{proposition}{Proposition}{Propositions}
\Crefname{proposition}{Proposition}{Propositions}
\crefname{lemma}{Lemma}{Lemmas}
\Crefname{lemma}{Lemma}{Lemmas}
\crefname{remark}{Remark}{Remarks}
\Crefname{remark}{Remark}{Remarks}
\crefname{definition}{Definition}{Definitions}
\Crefname{definition}{Definition}{Definitions}

\newcommand{\rd}{\mathrm{d}}

\usepackage[most]{tcolorbox}
\usepackage{xcolor}
\definecolor{lavender}{RGB}{240, 240, 250}   % hex #F1F1FA
\definecolor{mint}{RGB}{229,242,229}  % hex #F1F8F7

\definecolor{gray}{RGB}{246,246,246}  % hex #F1F8F7

\newtcolorbox{plainbox_mint}{
    enhanced,
    breakable,
    colback=mint,
    colframe=mint,
    boxrule=0pt,
    left=3pt,
    right=3pt,
    top=5pt,
    bottom=5pt,
    before skip=5pt,
    after skip=5pt
}

\newtcolorbox{plainbox_lavender}{
    enhanced,
    breakable,
    colback=lavender,
    colframe=lavender,
    boxrule=0pt,
    left=2pt,
    right=2pt,
    top=-8pt,
    bottom=0pt,
    before skip=5pt,
    after skip=5pt
}

\newtcolorbox{plainbox_gray}{
    enhanced,
    breakable,
    colback=gray,
    colframe=gray,
    boxrule=0pt,
    left=10pt,
    right=10pt,
    top=2pt,
    bottom=2pt,
    before skip=3pt,
    after skip=10pt
}

\title{Efficient Adjoint Matching for\\Fine-tuning Diffusion Models}

\author{%
  Jeongwoo Shin\thanks{Equal contribution.} \\
  Seoul National University \\
  \texttt{swswss@snu.ac.kr} \\
  \And
  Dongsoo Shin\footnotemark[1] \\
  Seoul National University \\
  \texttt{dongsoo@snu.ac.kr} \\
  \And
  Yuchen Zhu \\
  Georgia Institute of Technology \\
  \texttt{yzhu738@gatech.edu} \\
  \And
  Wei Guo \\
  Georgia Institute of Technology \\
  \texttt{wei.guo@gatech.edu} \\
  \And
  Yongxin Chen \\
  Georgia Institute of Technology \\
  \texttt{yongchen@gatech.edu} \\
  \And
  Joonseok Lee\thanks{Corresponding author.} \\
  Seoul National University,\\Hanwha Aerospace \\
  \texttt{joonseok@snu.ac.kr} \\
  \And
  Jaewoong Choi\footnotemark[2] \\
  Sungkyunkwan University \\
  \texttt{jaewoongchoi@skku.edu} \\
  \And
  Jaemoo Choi\footnotemark[2] \\
  Georgia Institute of Technology \\
  \texttt{jaemoo.choi@gatech.edu} \\
}

\begin{document}

\maketitle
\vspace{-1.7em}
{\centering
  \faGithub\ \href{https://github.com/swswss/EAM}{\texttt{\cfblue{github.com/swswss/EAM}}}\par}
\vspace{0.7em}
\begin{abstract}
    Reward fine-tuning has become a common approach for aligning pretrained diffusion and flow models with human preferences in text-to-image generation. Among reward-gradient-based methods, Adjoint Matching (AM) provides a principled formulation by casting reward fine-tuning as a stochastic optimal control (SOC) problem. However, AM inevitably requires a substantial computational cost: it requires (i) stochastic simulation of full generative trajectories under memoryless dynamics, resulting in a large number of function evaluations, and (ii) backward ODE simulation of the adjoint state along each sampled trajectory. In this work, we observe that both bottlenecks are closely tied to the \textit{non-trivial base drift} inherited from the pretrained model.
     Motivated by this observation, we propose \textbf{Efficient Adjoint Matching (EAM)}, which substantially improves training efficiency by reformulating the SOC problem with a \textit{linear base drift} and a correspondingly modified \textit{terminal cost}.
     This reformulation removes both sources of inefficiency; it enables training-time sampling with a few-step deterministic ODE solver and yields a closed-form adjoint solution that eliminates backward adjoint simulation.
    On standard text-to-image reward fine-tuning benchmarks, EAM converges up to 4× faster than AM and matches or surpasses it across various metrics including PickScore, ImageReward, HPSv2.1, CLIPScore and Aesthetics.
\end{abstract}

\section{Introduction}
\label{sec:intro}
% 1. FM, DM, T2I models. human preference misalignment $\rightarrow$ reward fine-tuning.
Diffusion~\cite{ho2020denoising,song2020score} and flow~\cite{lipman2022flow,liu2022flow,albergo2025stochastic} models have become the standard backbone for large-scale text-to-image (T2I) generation~\cite{rombach2022high,saharia2022photorealistic,esser2024scaling}. While their pretraining objective produces photorealistic samples, it is often poorly aligned with the qualities that actually matter in deployment such as aesthetic quality, prompt fidelity, and adherence to human preferences~\cite{xu2023imagereward,wu2023humanpreference}. To close this gap, reward-based fine-tuning has emerged as a standard recipe for aligning text-to-image models, where pretrained diffusion models are post-trained to maximize an external reward function that reflects these preferences~~\cite{liu2025flowgrpo, domingoenrich2025adjointmatching, zheng2026diffusionnft,choi2026rethinking,clark2024directly,xu2023imagereward,prabhudesai2024aligning}.

Among existing reward fine-tuning methods, reward-gradient-based methods provide direct supervision by leveraging the gradient of the reward function with respect to its input \cite{clark2024directly,xu2023imagereward,prabhudesai2024aligning,domingoenrich2025adjointmatching}. This gradient is readily available when the reward is parameterized by a differentiable network, as is typical for learned preference models such as CLIPScore, HPS, Aesthetic score, ImageReward, and PickScore~\cite{hessel2021clipscore,wu2023humanpreference,schuhmann2022aesthetics,xu2023imagereward,kirstain2023pickapic}.
% Among existing reward fine-tuning methods, \emph{reward-gradient-based} methods~\cite{clark2024directly,xu2023imagereward,prabhudesai2024aligning,domingoenrich2025adjointmatching} exploit the gradient of the reward with respect to its input, which is readily accessible whenever the reward is parameterized by a differentiable network~\cite{hessel2021clipscore,wu2023humanpreference,schuhmann2022aesthetics,xu2023imagereward,kirstain2023pickapic}, as is typical for learned preference models. 
While many reward fine-tuning methods use only scalar reward values~\cite{liu2025flowgrpo,zheng2026diffusionnft,choi2026rethinking}, reward-gradient-based methods additionally use the local gradient direction of the reward model. This provides each generated sample with direct supervision toward higher reward, offering an effective and sample-efficient route when reliable reward gradients are available.
Adjoint Matching (AM)~\cite{domingoenrich2025adjointmatching} introduces a theoretically principled framework for reward-gradient-based fine-tuning. Specifically, AM casts reward fine-tuning as a stochastic optimal control (SOC) problem and learns an additional control that transforms the pretrained model toward the reward-tilted target distribution $p^\star_1$, defined as
\begin{align}\label{eq:soc_optimal_p1}
    p^\star_1 (x) \propto e^{\beta r(x)} p_\text{data} (x),
\end{align}
where $r(x)$ is a given reward function, $\beta>0$ is the constant that controls the strength of reward guidance, and $p_\text{data}$ denotes the data distribution.
However, AM requires heavy computational cost for optimization (\cref{sec:method:AM_inefficiency}). First, AM relies on \textbf{forward SDE simulation} of full generative trajectories under memoryless dynamics. This SDE simulation requires a large number of function evaluations (NFEs), substantially increasing sampling time during training. 
% Moreover, this training trajectory differs from standard evaluation, where samples are typically generated using a few-step deterministic ODE solver. 
Second, AM requires \textbf{backward ODE simulation} for adjoint state along sampled trajectory to propagate reward-gradient signal (\cref{sec:prelim:am}). Together, these two simulation procedures dominate the training cost.

Our key observation is that these bottlenecks arise from the choice of the drift term of the base dynamics (\cref{sec:method:linear_drift}). Building on this observation, we redesign the base dynamics using a \textit{simple linear drift} that satisfies the desired conditions. We accordingly modify the \textit{terminal cost} in the SOC problem so that the optimal terminal distribution matches with $p^\star_1$ in \cref{eq:soc_optimal_p1}. This reformulation leads to our algorithm, \textbf{Efficient Adjoint Matching (EAM)}, which removes both costly simulations in AM in the following way. First, forward trajectory simulation becomes solver-agnostic: the endpoint image $X_1$ can be generated using any efficient few-step deterministic ODE solver~\cite{zhang2022deis,zhang2023gddim}, and intermediate states $X_t$ can be sampled from the original noising kernel $q_t(\cdot\mid X_1)$. Second, the backward ODE simulation for computing the adjoint state could be replaced by closed-form solution, eliminating backward ODE simulation entirely.
Despite this change of base dynamics, we prove that EAM shares the same optimal path measure as AM, so the efficiency gain comes without altering the optimum.
Empirically, EAM reduces the per-iteration training cost by up to 4× compared to AM, while matching or surpassing its performance across various human preference metrics. Our contributions can be summarized as follows:
% Our key observation is that these bottlenecks arise from the choice of the drift term in the base dynamics (\cref{sec:method:linear_drift}). Building on this, we redesign the base dynamic to have a simple linear drift so that both costly steps disappear at once (\cref{sec:method:reform}): (i) Trajectory simulation becomes solver-agnostic, so that we can use any few-step ODE solver. The image $X_1$ can be obtained with an efficient ODE in only a few steps and intermediate state $X_t$ can be obtained by simply noising this image. (ii) backward ODE simulation for adjoint state reduces to one-step closed-form calculation, removing the simulation entirely. \jeongwoo{Since changing the base dynamic also changes the uncontrolled terminal distribution, we rederive the terminal cost to ensure that the optimal control remains targeted at the desired reward-tilted distribution. (\cref{sec:method:reform}).}

\begin{figure}[t]
    \centering
    \includegraphics[width=1.\linewidth]{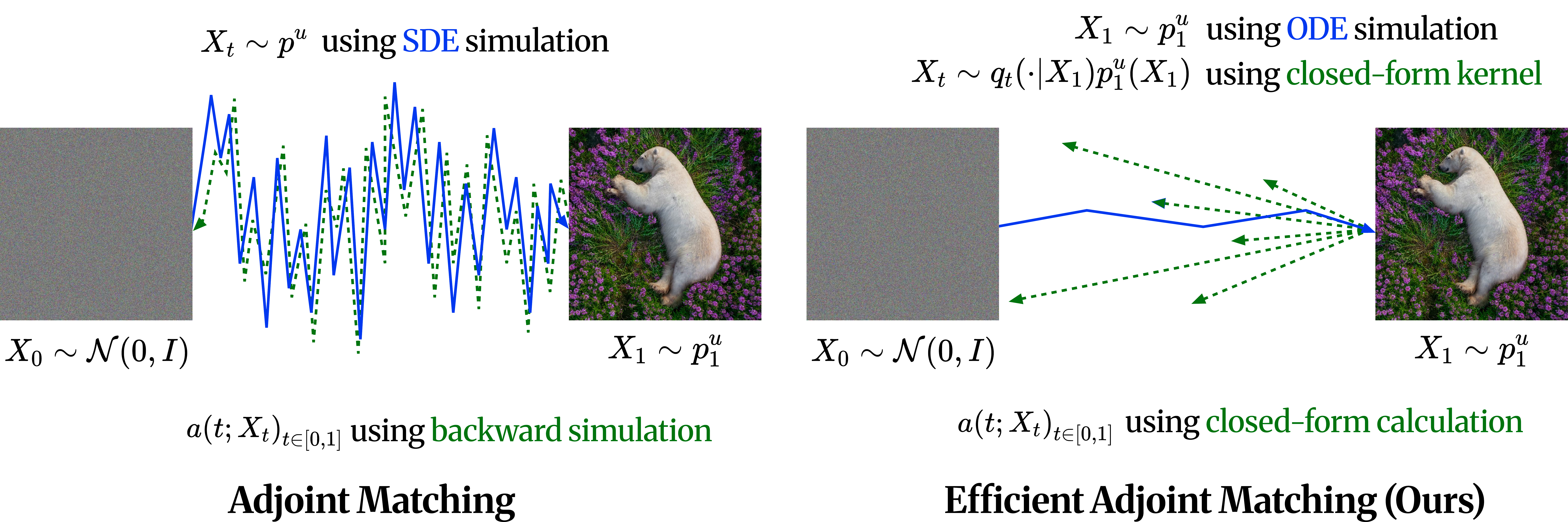}
    \vspace{-.4cm}
    \caption{
        \textbf{Comparison of Efficient Adjoint Matching (EAM) with Adjoint Matching (AM).} (\emph{Left}) AM relies on a stochastic SDE solver to construct each training trajectory and a sequential backward simulation to obtain the adjoint state along that trajectory. (\emph{Right}) EAM eliminates both: intermediate states $X_t$ are obtained by first simulating the endpoint $X_1$ with a few-step ODE and then sampling $X_t$ from the original noising kernel $q_t(\cdot| X_1)$, while the adjoint state is given by a single closed-form evaluation, removing the backward simulation entirely.
    }
    \label{fig:intro}
    \vspace{-.4em}
\end{figure}
% Contributions: 1) Reformulation of base dynamic in SOC based reward tilting framework to accelerate the learning process, while keeping the theoretical completion of AM. 2) Demonstration through the experiments showing $4\times$ faster convergence.
\begin{itemize}
    \item We propose \textbf{Efficient Adjoint Matching (EAM)}, an efficient reward-gradient-based fine-tuning algorithm derived by redesigning the \textit{base drift} and a \textit{terminal cost} of SOC problem. 
    \item We \textbf{characterize the linear base drifts} satisfying our design requirements, enabling efficient ODE-based trajectory construction and a closed-form adjoint state, and prove that the resulting \textbf{optimal path measure coincides with that of AM}.
    \item We show that EAM achieves \textbf{comparable or better performance} on standard text-to-image reward fine-tuning benchmarks while converging up to \textbf{4× faster} than AM.
\end{itemize}
% \textbf{Contributions.} Our contributions are three-fold:
% \vspace{-0.5em}
% \begin{itemize}[leftmargin=*]
%     \item We propose \textbf{Efficient Adjoint Matching}, a highly efficient reward fine-tuning method that \textbf{reformulates the base dynamic} so that trajectory construction uses an \textbf{efficient deterministic solver} and the matching target reduces to a \textbf{closed-form, one-step calculation}.
%     \item We \textbf{characterize the family of linear base drifts} which is \textbf{unique} under our requirements, enabling both efficient ODE-based trajectory sampling and a closed-form adjoint state.
%     \item On standard text-to-image reward fine-tuning benchmarks, we empirically demonstrate that our method achieves \textbf{comparable or even better results} among various metrics with up to \textbf{$\mathbf{4\times}$ faster convergence} than Adjoint Matching.
% \end{itemize}
% 뭔가 prove한 내용? 어떤 property 만족하는 뭔가를 보였다, property의 characterization, 어떤 base drift 찾음
\section{Preliminaries}
\label{sec:preliminaries}

\subsection{Diffusion and Flow Models}
Diffusion~\cite{ho2020denoising,song2020score} and flow matching~\cite{lipman2022flow,albergo2025stochastic,liu2022flow} models share a common formulation through a pretrained velocity field 
$v^{\mathrm{pt}}:\mathbb{R}^d\times[0,1]\rightarrow\mathbb{R}^d$, learned by velocity matching along a conditional probability path. 
Throughout this paper, we focus on the linear path~\cite{lipman2022flow,liu2022flow}. 
Equivalently, given an endpoint $X_1$, intermediate states are sampled from the following noising kernel:
\begin{align}
    q_t(X_t\mid X_1)
    :=
    \mathcal{N}\!\left(tX_1,(1-t)^2 I\right),
    \qquad
    \emph{i.e.,}\quad
    X_t=tX_1+(1-t)\epsilon,\;\; \epsilon\sim\mathcal{N}(0,I),
    \label{eq:flow_kernel}
\end{align}
where $X_1\sim p_{\mathrm{data}}$. 
We denote the endpoint distributions by $p_0=\mathcal{N}(0,I)$ and $p_1=p_{\mathrm{data}}$. 
The pretrained velocity field $v^{\mathrm{pt}}$ induces the following forward generative dynamic:
\begin{align}
    \rd X_t
    =
    \left(
    -\frac{1}{t}X_t+2v^{\mathrm{pt}}(X_t,t)
    \right)\rd t
    +
    \sigma(t)\,\rd W_t,
    \qquad
    X_0\sim p_0,
    \qquad
    \sigma(t)=\sqrt{\frac{2(1-t)}{t}}.
    \label{eq:flow_gen_dynamic}
\end{align}
We write $p^{\mathrm{pt}}$ for the path measure induced by~\cref{eq:flow_gen_dynamic}, and throughout the paper we fix the diffusion coefficient $\sigma(t)$ as in~\cref{eq:flow_gen_dynamic}. 
Since the pretrained model is trained to approximate the data distribution, we use the standard idealization $p^{\mathrm{pt}}_1=p_{\mathrm{data}}$.

\subsection{Stochastic Optimal Control for Fine-tuning Diffusion Models}

\textbf{Problem Setting and Notations.}
We consider the following SOC problem:
\begin{align}
    \min_u\;&
    \mathbb{E}_{p^u}\!\left[
    \int_0^1
    \frac{1}{2}\,\|u(X_t,t)\|^2\,\rd t
    + g(X_1)
    \right],
    \label{eq:control-objective}
    \\
    \text{s.t.}\quad
    \rd X_t
    &=
    \bigl(
    b(X_t,t) + \sigma(t)\,u(X_t,t)
    \bigr)\,\rd t
    + \sigma(t)\,\rd W_t,
    \qquad
    X_0 \sim p_0,
    \label{eq:controlled-sde}
\end{align}
where $X_t\in\mathbb{R}^d$ is the state of the controlled dynamic~\eqref{eq:controlled-sde}, $u:\mathbb{R}^d\times[0,1]\rightarrow\mathbb{R}^d$ is the control vector field, $b:\mathbb{R}^d\times[0,1]\rightarrow\mathbb{R}^d$ is the base drift, 
% $\sigma:[0,1]\rightarrow\mathbb{R}$ is the diffusion coefficient, 
and $g:\mathbb{R}^d\rightarrow\mathbb{R}$ is the terminal cost. We denote by $p^u$ the path measure induced by the controlled dynamic~\eqref{eq:controlled-sde}, and by $p^{\text{base}}$ the path measure induced by the uncontrolled base dynamic, \emph{i.e.}, \cref{eq:controlled-sde} with $u\equiv 0$. With a properly designed terminal cost $g$, the controlled dynamic~\eqref{eq:controlled-sde} under the optimal control $u^\star$ reaches the target distribution $p^\star_1$ at $t=1$.

\begin{plainbox_mint}
\textbf{\textbf{Standard setup for SOC-based reward fine-tuning.}}
Suppose the base drift $b(x,t)$ and the terminal cost $g(x)$ are defined as follows:
\begin{equation}\label{eq:reward_soc_setting}
    b(x,t) = -\frac{1}{t}x + 2v^{\mathrm{pt}}(x,t),
    \qquad
    g(x) = -\beta r(x).
\end{equation}
Then, the optimal solution $p^{u^\star}$ for the SOC problem yields $p^{u^\star}_1(x)=p^\star_1(x) \propto e^{\beta r(x)} p_\text{data}(x)$. 
\end{plainbox_mint}
% with $\sigma(t)$ fixed as in~\cref{eq:flow_gen_dynamic}. 

% by definition, $p^{\mathrm{base}}=p^{\mathrm{pt}}$, and solving the SOC problem yields $p^{u^\star}_1=p^\star_1$. 
Furthermore, to yield an unbiased estimator of $u^\star$, the base dynamic must be memoryless~\cite{domingoenrich2025adjointmatching}:
\begin{align}
    p^{\mathrm{base}}_{0,1}(X_0,X_1)
    =
    p^{\mathrm{base}}_0(X_0)\,p^{\mathrm{base}}_1(X_1).
    \label{eq:memoryless_cond}
\end{align}
The standard choice~\eqref{eq:reward_soc_setting} satisfies this condition.
% \end{remark}

% By taking the base dynamic in \cref{eq:controlled-sde} to be the pretrained generative dynamic~\eqref{eq:flow_gen_dynamic} (\emph{i.e.}, $p^{\text{base}}=p^{\text{pt}}$), and given a differentiable reward function $r:\mathbb{R}^d\rightarrow\mathbb{R}$ with a scaling constant $\beta>0$, the choice of terminal cost $g(x)=-\beta r(x)$ yields the reward-tilted terminal distribution $p^{u^\star}_1(x)\propto p^{\text{pt}}_1(x)\exp(\beta r(x))$ at optimality~\cite{domingoenrich2025adjointmatching}.

% \textbf{Memoryless condition.}
% To ensure that this regression yields an unbiased estimate of the optimal control $u^\star$, the base dynamic must satisfy the \emph{memoryless condition}~\cite{domingoenrich2025adjointmatching}:
% \begin{align}
%     p^{\text{base}}_{0,1}(X_0,X_1)\;\overset{\text{memoryless}}{=}\;p^{\text{base}}_0(X_0)\,p^{\text{base}}_1(X_1).
%     \label{eq:memoryless_cond}
% \end{align}
% Pretrained generative dynamic~\eqref{eq:flow_gen_dynamic} satisfies this condition.
% By solving this problem, we aim to have our controlled process $X_t$ end up at the reward-tilted distribution at its terminal state $X_1$. Corresponding uncontrolled base process for this SOC problem would be 
% \begin{align}
%     \text dX_t
%     &=
%     b(X_t,t)
%     \,\text dt
%     + \sigma(t)\,\text d \tilde B_t,
%     \qquad
%     X_0 \sim p_0 .
% \end{align}

%%%%%%%%%%%%%%%%%%%%%%%%%%%%%%%%%%%%%%%%%%%%%%%%%%%%%%%%%%%%
%%%%%%%%%%%%%%%%%%%%%%%%%%%%%%%%%%%%%%%%%%%%%%%%%%%%%%%%%%%%

\subsection{Adjoint Matching}
\label{sec:prelim:am}
Adjoint Matching (AM)~\cite{domingoenrich2025adjointmatching} provides an efficient way to solve the SOC problem~\eqref{eq:control-objective} under \cref{eq:reward_soc_setting} by learning the control $u$ through regression onto the lean adjoint state $a:\mathbb{R}^d\times[0,1]\rightarrow\mathbb{R}^d$:
\begin{plainbox_lavender}
\begin{align}
    \mathcal{L}_{\mathrm{AM}}
    &:=
    \frac{1}{2}
    \int_0^1
    \left\|
    u(X_t,t) + \sigma(t)\,a(t;X_t)
    \right\|^2 \,\rd t,
    \qquad X_t\sim p^{\bar u},
    \quad \bar u = \operatorname{stopgrad}(u),
    \label{eq:am-loss}
    \\
    \text{where}\qquad 
    &\frac{\rd}{\rd t} a(t;X_t)
    =
    -\nabla_x b(X_t,t)^\top a(t;X_t),
    \qquad  a(1;X_1) = \nabla g(X_1).
    \label{eq:adj-ode}
\end{align}
\end{plainbox_lavender}

% \jeongwoo{\textbf{Optimization procedure.}}
% (iv) regression 계산을 한다를 4번으로
% ode integration -> backward simulation
Each training iteration of AM proceeds in four steps: (i) simulate a trajectory $\{X_t\}_{t\in[0,1]}\sim p^{\bar u}$ of the controlled dynamic~\eqref{eq:controlled-sde} with a stochastic solver, as required by the memoryless condition~\eqref{eq:memoryless_cond}; (ii) compute the gradient of the terminal cost $\nabla g(X_1) = -\beta\nabla r(X_1)$; (iii) simulate the adjoint ODE backward along the stored trajectory to obtain the adjoint states~\eqref{eq:adj-ode}; (iv) and regress the control to the matching target~\eqref{eq:am-loss}.
% to obtain the regression target in \cref{eq:am-loss}.
% a stochastic simulation of the controlled dynamic~\eqref{eq:controlled-sde} to satisfy memoryless condition~\eqref{eq:memoryless_cond} to store a trajectory $\{X_t\}_{t\in[0,1]}\sim p^{\bar u}$, and (ii) a backward integration of the adjoint ODE~\eqref{eq:adj-ode} along this trajectory to obtain the regression target in \cref{eq:am-loss}.

% When the base dynamic is taken to be the pretrained generative dynamic~\eqref{eq:flow_gen_dynamic}, the memoryless condition~\eqref{eq:memoryless_cond} is satisfied at $\lambda_s=1$, in which case $p^{\text{base}}_1=p^{\text{pt}}_1$ and the reward-tilted target reduces to $p^{u^\star}_1(x)\propto p^{\text{pt}}_1(x)\exp(\beta r(x))$, where $p^{\text{pt}}_1$ denotes the terminal distribution of the pretrained model~\eqref{eq:flow_gen_dynamic}. 

%%%%%%%%%%%%%%%%%%%%%%%%%%%%%%%%%%%%%%%%%%%%%%%%%%%%%%%%%%%%
%%%%%%%%%%%%%%%%%%%%%%%%%%%%%%%%%%%%%%%%%%%%%%%%%%%%%%%%%%%%

\section{Efficient Adjoint Matching}
% AM으로 바로해도 무관
\subsection{Inefficiency of Adjoint Matching}
\label{sec:method:AM_inefficiency}
Although Adjoint Matching (AM)~\cite{domingoenrich2025adjointmatching} achieves strong empirical alignment performance among reward gradient-based fine-tuning methods with distributional guarantees, its per-iteration cost has not been carefully examined. We identify two computational bottlenecks:
\begin{itemize}[leftmargin=2em]
    \item \textbf{Forward SDE simulation.} The regression in \cref{eq:am-loss} requires sampling $X_t\sim p^{\bar u}$ under memoryless condition~\eqref{eq:memoryless_cond}. It typically requires substantially more sampling steps than its deterministic counterpart and additionally requires storing the full trajectory in memory for the subsequent adjoint pass.
    \item \textbf{Backward adjoint simulation with Jacobian--Vector Product (JVP).} Computing the lean adjoint state~\eqref{eq:adj-ode} requires a backward simulation along the same trajectory, with each step evaluating the JVP $\nabla_x b(X_t,t)^\top a(t;X_t)$. Since the base drift $b$ contains the pretrained velocity $v^{\text{pt}}$, this JVP must be propagated through the network at every integration step.
\end{itemize}

% base drift 바꿔줄거고
% reward-tilted p (eq) 만족하기 위해 g까지 바꿔줘야한다
% 3.2, 3.3 가이던스
Our core observation is that both inefficiencies stem from the \textbf{base drift} $b$ in \cref{eq:reward_soc_setting}. In \cref{sec:method:linear_drift}, we first characterize the properties required for an efficient base dynamic, namely \textbf{simulation-free adjoint computation} and \textbf{efficient trajectory construction}. Then, in \cref{sec:method:reform}, we instantiate these properties by \textbf{redesigning the base drift} and \textbf{rederiving the terminal cost} $g$, so that the optimal controlled terminal marginal remains the reward-tilted distribution in~\cref{eq:soc_optimal_p1}.

% linear + memoryless
% gray box like rethinking rl
% linear -> remark adj, linear+mem -> remark xt|x1
\subsection{Characterizing an Efficient Base Dynamic}
\label{sec:method:linear_drift}
% We first identify two structural conditions for an efficient base dynamic:
% a linear base drift enables closed-form adjoint computation, while memorylessness enables efficient construction of intermediate states.
We identify two structural conditions for an efficient base dynamic: linearity of the base drift, which enables closed-form adjoint computation, and memorylessness, which enables efficient construction of intermediate states.
% \jeongwoo{, whose base path measure we denote by $p^{\mathrm{base},\mathrm{EAM}}$.} 
% We instantiate these conditions with our redesigned base dynamic in~\cref{sec:method:reform}.
\vspace{0.2em}
\begin{plainbox_gray}
\begin{itemize}[label={}, itemsep=3pt, topsep=0pt, parsep=0pt, partopsep=0pt, leftmargin=*]
    \item \textbf{Condition 1.} The base drift $b$ is \textbf{linear} in $x$.
    \item \textbf{Condition 2.} The base dynamic is \textbf{memoryless}~\eqref{eq:memoryless_cond}.
\end{itemize}
\end{plainbox_gray}
\vspace{-0.6em}

% \begin{plainbox_gray}
% \textbf{Condition 1.} The base drift $b$ is \textbf{linear} in $x$.\\
% \textbf{Condition 2.} The base dynamic is \textbf{memoryless}~\eqref{eq:memoryless_cond}.
% \end{plainbox_gray}
We now show that these two properties jointly eliminate both bottlenecks identified in~\cref{sec:method:AM_inefficiency}.

\textbf{Condition 1: Linear drift enables a closed-form adjoint.} 
Suppose the base drift is linear in $x$, i.e., $b(x,t)=D(t)x$ for some scalar function $D:[0,1]\rightarrow\mathbb{R}$. 
Then the lean adjoint state defined by~\cref{eq:adj-ode} admits the closed-form solution
\begin{align}
    a(t;X_t)=\exp\!\left(\int_t^1 D(\tau)\,\rd\tau\right) a(1;X_1).
    \label{eq:closed_adj}
\end{align}
Since $\nabla_x b(x,t)=D(t)I$, the JVP in~\cref{eq:adj-ode} reduces to a scalar multiplication and no longer requires differentiating through the pretrained network $v^{\text{pt}}$. 
Thus, \cref{eq:closed_adj} eliminates the backward adjoint ODE simulation: the adjoint state at any time $t$ is the endpoint gradient $a(1;X_1)=\nabla g(X_1)$ scaled by a coefficient that depends only on $t$. 
Accordingly, the AM loss in~\cref{eq:am-loss} reduces to
\begin{align}
    % \mathcal{L}_{\mathrm{AM}}
    % =
    \mathbb{E}_{(X_t,X_1)\sim p^{\bar u}}\left[
    \frac{1}{2}
    \int_0^1
    \left\|
    u(X_t,t) + \sigma(t)\exp\!\left(\int_t^1 D(\tau)\,\rd\tau\right)\nabla g(X_1)
    \right\|^2 \,\rd t\right].
    \label{eq:am-loss-closed-adj}
\end{align}

\textbf{Condition 2: Memorylessness enables direct sampling of $X_t$ from $X_1$.}
Even with the closed-form adjoint in~\cref{eq:am-loss-closed-adj}, the loss still requires samples of the joint pair $(X_t,X_1)$.
A direct implementation obtains this pair by simulating the full stochastic trajectory.
When the base dynamics are memoryless~\eqref{eq:memoryless_cond}, this full rollout can be bypassed \citep{guo2025proximal}: under SOC optimality, the intermediate state can be sampled directly from the endpoint $X_1$,
\begin{align}
    \int p^{\mathrm{base}}_{t|0,1}(X_t\mid X_0,X_1)\,p^{u^\star}(X_0,X_1)\rd X_0
    =
    p^{\mathrm{base}}_{t|1}(X_t\mid X_1)\,p^{u^\star}(X_1).
    \label{eq:reciprocal}
\end{align}
Thus, instead of simulating the full path, we first sample $X_1\sim p^{\bar u}_1$ using an efficient ODE solver, and then sample $X_t$ from the conditional base kernel $p^{\mathrm{base}}_{t|1}(X_t\mid X_1)$.
Moreover, since the base drift is linear, this conditional distribution is available in closed form, so each intermediate state can be obtained by adding the appropriate amount of forward noise to the endpoint $X_1$ in a single step.
The control matching loss therefore reduces further to
\begin{plainbox_lavender}
\begin{align}
    \mathcal{L}_{\mathrm{EAM}}
    =
    \mathbb{E}_t \mathbb{E}_{X_1\sim p^{\bar u}_1,\, X_t\sim p^{\mathrm{base}}_{t|1}(\cdot\mid X_1)}
    \left[
    \frac{1}{2} \left\|
    u(X_t,t) + \sigma(t)e^{\left(\int_t^1 D(\tau)\,\rd\tau\right)} \nabla g(X_1)
    \right\|^2 \right].
    \label{eq:am-loss-closed-adj-kernel}
\end{align}
\end{plainbox_lavender}

Since~\cref{eq:reciprocal} holds only at optimality, this sampling is approximate during training, while the regression target remains exact for any control. When $\bar u=u^\star$, the sampling becomes exact and $\mathcal{L}_{\mathrm{EAM}}$ reduces to the AM regression~\eqref{eq:am-loss-closed-adj}, whose minimizer is $u^\star$~\cite{domingoenrich2025adjointmatching}. Hence $u^\star$ is a fixed point of EAM, although convergence is not guaranteed in general. This type of fixed-point scheme is widely adopted in recent SOC-based methods~\cite{havens2025adjointsampling,liu2025adjointschrodingerbridgesampler,havens2026flowsampling}.

\textbf{Summary.}
\textbf{Condition 1} requires the base drift to be linear, which yields a closed-form adjoint and eliminates backward adjoint simulation.
\textbf{Condition 2} requires the base dynamic to be memoryless, which enables endpoint-conditioned noising in place of full SDE simulation.
Together, these conditions remove the two computational bottlenecks identified in~\cref{sec:method:AM_inefficiency}.
We next instantiate them by redesigning the base drift and rederiving the terminal cost to preserve 
the reward-tilted target~\eqref{eq:soc_optimal_p1}.
% \clearpage

%\input{table/am_ours_comparison}
\newcommand{\cmark}{\textcolor{green!60!black}{\ding{51}}}
\newcommand{\xmark}{\textcolor{red!70!black}{\ding{55}}}

\begin{table}[t]
\centering
\caption{Comparison between AM and our EAM.}
\label{tab:am-vs-ours}
\renewcommand{\arraystretch}{1.4}
\resizebox{\linewidth}{!}{%
\begin{tabular}{l|cc}
\toprule
\textbf{Method} & \textbf{Adjoint Matching}~\cite{domingoenrich2025adjointmatching} & \textbf{Efficient Adjoint Matching} \\
\midrule
\multirow{2}{*}{Control problem} &  \multicolumn{2}{c}{ $\min\limits_u \mathbb{E}_{p^u}\!\left[\int_0^1 \tfrac{1}{2}\|u(X^u_t,t)\|^2\,\rd t + g(X^u_1)\right]$} \\
 & \multicolumn{2}{c}{ s.t. $\rd X^u_t = \big(b(X^u_t,t) + \sigma(t)\,u(X^u_t,t)\big)\rd t + \sigma(t)\,\rd W_t,\quad X_0^u \sim p_0$} \\
\midrule
Base drift $b(x,t)$ & \cellcolor{mint} $-\tfrac{1}{t}\,x + 2\,v^{\text{pt}}(x,t)$ & \cellcolor{mint} $\dfrac{2Ct^2-1}{t\,(2Ct^2-2t+1)}\,x$ \\
% \midrule
% Diffusion $\sigma(t)$ & \multicolumn{2}{c}{$\sqrt{2(1-t)/t}$} \\
\midrule
Grad. of terminal cost 
$\nabla g(x)$ & \cellcolor{mint} $-\beta \nabla r(x)$ & \cellcolor{mint} $-\frac{x}{2C-1} -\nabla\log p^{\text{pt}}_1(x) -\beta\nabla r(x)$ \\
\midrule
Loss objective & \cellcolor{lavender} 
\cref{eq:am-loss} & \cellcolor{lavender} 
\cref{eq:am-loss-closed-adj-kernel}
\\
\midrule
Adjoint simulation-free &  \xmark & \cmark \\
ODE solver for $\{X_t\}_{t\in[0,1]}$ &  \xmark &   \cmark \\
No storage of $\{X_t\}_{t\in[0,1]}$ &  \xmark &   \cmark \\
\bottomrule
\end{tabular}%
}
\end{table}

\subsection{Redesigning the Base Drift and Terminal Cost for Efficient Adjoint Matching}
\label{sec:method:reform}
We now instantiate the conditions identified in~\cref{sec:method:linear_drift}. 
Since fine-tuning starts from the pretrained generative dynamic, diffusion coefficient $\sigma(t)$ is fixed as in \cref{eq:flow_gen_dynamic}. 
Under this restriction, \textbf{Conditions 1 \& 2} alone do not uniquely determine the base dynamic. 
% We therefore impose one additional requirement: 
% the conditional distribution used to sample an intermediate state $X_t$ from the endpoint $X_1$, i.e. $p^{\mathrm{base}}(X_t\mid X_1)$, should match the original noising kernel $q_t$ defined in~\cref{eq:flow_kernel}:
We further narrow down the design space by requiring the base noising kernel $p^{\mathrm{base}}_{t|1}(\cdot | X_1)$ to match the original noising kernel $q_t$~\eqref{eq:flow_kernel}:
% We therefore impose one additional requirement: the noising kernel of the base dynamic $p^{\text{\jeongwoo{base}}}(X_t|X_1)$ matches the original perturbation kernel in~\cref{eq:flow_kernel}.

\vspace{0.6em}
\begin{plainbox_gray}
\begin{itemize}[label={}, itemsep=0pt, topsep=0pt, parsep=0pt, partopsep=0pt, leftmargin=*]
    \item \textbf{Condition 3.} The base noising kernel matches the original noising kernel in \cref{eq:flow_kernel}:
    \begin{align}
        p^{\mathrm{base}}_{t|1}(\cdot\mid X_1)
        =
        q_t(\cdot\mid X_1)
        =
        \mathcal{N}\!\left(tX_1,(1-t)^2I\right).
        \label{eq:condition_kernel_match}
    \end{align}
\end{itemize}
% \end{mdframed}
\end{plainbox_gray}
\vspace{-0.6em}

\begin{algorithm}[H]
\caption{Efficient Adjoint Matching}
\label{alg:ours}
\begin{algorithmic}[1]
\Require Pretrained velocity model $v^{\text{pt}}$, reward $r(x)$, LoRA parameters $\theta$
\State Initialize $v^{\text{ft}} \leftarrow \text{LoRA}_\theta(v^{\text{pt}})$
\Repeat
    \State Sample $X_1$ via ODE simulation: 
    \begin{equation}
    \rd X_t = \bar{v}^{\text{ft}}(X_t,t)\,\rd t,\quad X_0 \sim p_0, \quad \bar v^{\text{ft}} = \operatorname{stopgrad}(v^{\text{ft}})
    \label{eq:alg:sampling}
    \end{equation}
    \State Construct intermediate state using 
    % \jeongwoo{$p^{\mathrm{base,EAM}}_{t|1}$}:
    $q_t(\cdot \mid X_1)$~\eqref{eq:flow_kernel}:
    \begin{equation}
        X_t = tX_1 + (1-t)\epsilon,\quad \epsilon \sim \mathcal{N}(0, I), \quad t\sim \mathcal U[0,1]
    \label{eq:alg:xt}
    \end{equation}
    \State Compute $\nabla g(X_1)$~\eqref{eq:ours_g} using \cref{eq:score_est}
    \State Minimize the loss objective $\mathcal{L}_{\text{EAM}}$ in \cref{eq:am-loss-closed-adj-kernel}
    % \begin{equation}
    %     \mathcal{L}(\theta)
    %     = \mathbb{E}_{X_t, X_1}\!\left[
    %     \frac{1}{\sigma(t)^2}\left\|
    %     v^{\text{ft}}(X_t,t)
    %     +\frac{1 - 2Ct}{2Ct^2 - 2t + 1}X_t
    %     + \frac{(2C-1)(1-t)}{2Ct^2 - 2t + 1}\nabla g(X_1)
    %     \right\|^2
    %     \right]
    % \label{eq:alg:loss}
    % \end{equation}
    \State Update $\theta$ by gradient descent on $\nabla_\theta \mathcal{L}_{\text{EAM}}(\theta)$
\Until{convergence}
\State \Return $v^{\text{ft}}$
\end{algorithmic}
\end{algorithm}
\vspace{-0.5em}
The following proposition characterizes the resulting admissible family of base drifts.

\begin{plainbox_mint}
\begin{proposition}[Family of admissible linear base drifts]
\label{prop:reformed_linear_drift}
Assume that the diffusion coefficient $\sigma(t)$ is fixed as in~\cref{eq:flow_gen_dynamic}. 
Then the base dynamic satisfies \textbf{Conditions 1--3} if and only if its drift is
\begin{align}
    b(x,t)=D(t)x,
    \qquad
    D(t)
    =
    \frac{2Ct^2-1}{t\,(2Ct^2-2t+1)},
    \qquad C>\frac{1}{2}.
    \label{eq:reformed_base_drift}
\end{align}
\end{proposition}
\end{plainbox_mint}

Since the base dynamic is now redesigned, the standard terminal cost~\eqref{eq:reward_soc_setting} no longer yields the desired reward-tilted distribution in~\cref{eq:soc_optimal_p1}. 
We therefore redesign the terminal cost $g(x)$ to account for our new base dynamic induced by \cref{eq:reformed_base_drift}.
\begin{plainbox_mint}
\begin{proposition}[Terminal cost correction]
\label{prop:terminal_cost}
Let the base dynamic in~\cref{eq:controlled-sde} use the redesigned drift in~\cref{eq:reformed_base_drift}, whose terminal marginal is $p^{\mathrm{base}}_1=\mathcal{N}(0,(2C-1)I)$. 
If
\begin{align}
    \nabla g(x)
    =
    \nabla\log p^{\mathrm{base}}_1(x)
    -
    \nabla\log p^{\mathrm{pt}}_1(x)
    -
    \beta\nabla r(x),
    \label{eq:ours_g}
\end{align}
then the optimal controlled terminal distribution satisfies $p^{u^\star}_1=p^\star_1$, with $p^\star_1$ defined in~\cref{eq:soc_optimal_p1}.
\end{proposition}
\end{plainbox_mint}

Proposition~\ref{prop:terminal_cost} guarantees that EAM attains the reward-tilted terminal distribution. We further show that the entire optimal path measure coincides with that of AM.

\begin{plainbox_mint}
\begin{theorem}[Equivalence of optimal path measures]
\label{thm:path_equiv}
Let $u^\star_{\mathrm{AM}}$ and $u^\star_{\mathrm{EAM}}$ be the optimal controls of the SOC problem~\eqref{eq:control-objective}--\eqref{eq:controlled-sde} under the AM setting~\eqref{eq:reward_soc_setting} and under our setting~\eqref{eq:reformed_base_drift}--\eqref{eq:ours_g}, respectively. Under standard well-posedness assumptions, their induced path measures coincide:
\begin{align}
    p^{u^\star_{\mathrm{AM}}} = p^{u^\star_{\mathrm{EAM}}}.
    \label{eq:path_equiv}
\end{align}
In particular, both optimal processes share the drift $b^\star(x,t) = \frac{x}{t} + \sigma^2(t)\nabla\log p^\star_t(x)$, where $p^\star_t(x) = \int q_t(x\mid x_1)\,p^\star_1(x_1)\,\rd x_1$.
\end{theorem}
\end{plainbox_mint}

Note that the optimal controls themselves differ ($u^\star_{\mathrm{AM}} \neq u^\star_{\mathrm{EAM}}$), since they are measured relative to different base drifts. Consequently, the control costs and the optimization landscapes differ, which may lead to different training behavior under finite model capacity. Their optimal controlled processes, however, are identical. Hence EAM is not an approximation of AM: it reaches exactly the same optimum through a different reference process. The proof is provided in~\cref{appendix:proof:path_equiv}.

\begin{remark}[Instantiating the EAM objective]
Our EAM can be instantiated by plugging our redesigned base drift~\eqref{eq:reformed_base_drift} and corrected terminal cost~\eqref{eq:ours_g} into our loss objective $\mathcal{L}_{\text{EAM}}$~\eqref{eq:am-loss-closed-adj-kernel}. 
\end{remark}
The remaining practical questions are how to approximate $\nabla\log p^{\mathrm{pt}}_1$ in terminal cost~\eqref{eq:ours_g} and how to parameterize control $u(x,t)$ in our loss objective~\eqref{eq:am-loss-closed-adj-kernel}; we address both in the next section.

\begin{figure}[t]
    \centering
    % \vspace{-0.2cm}
    \includegraphics[width=1.\linewidth]{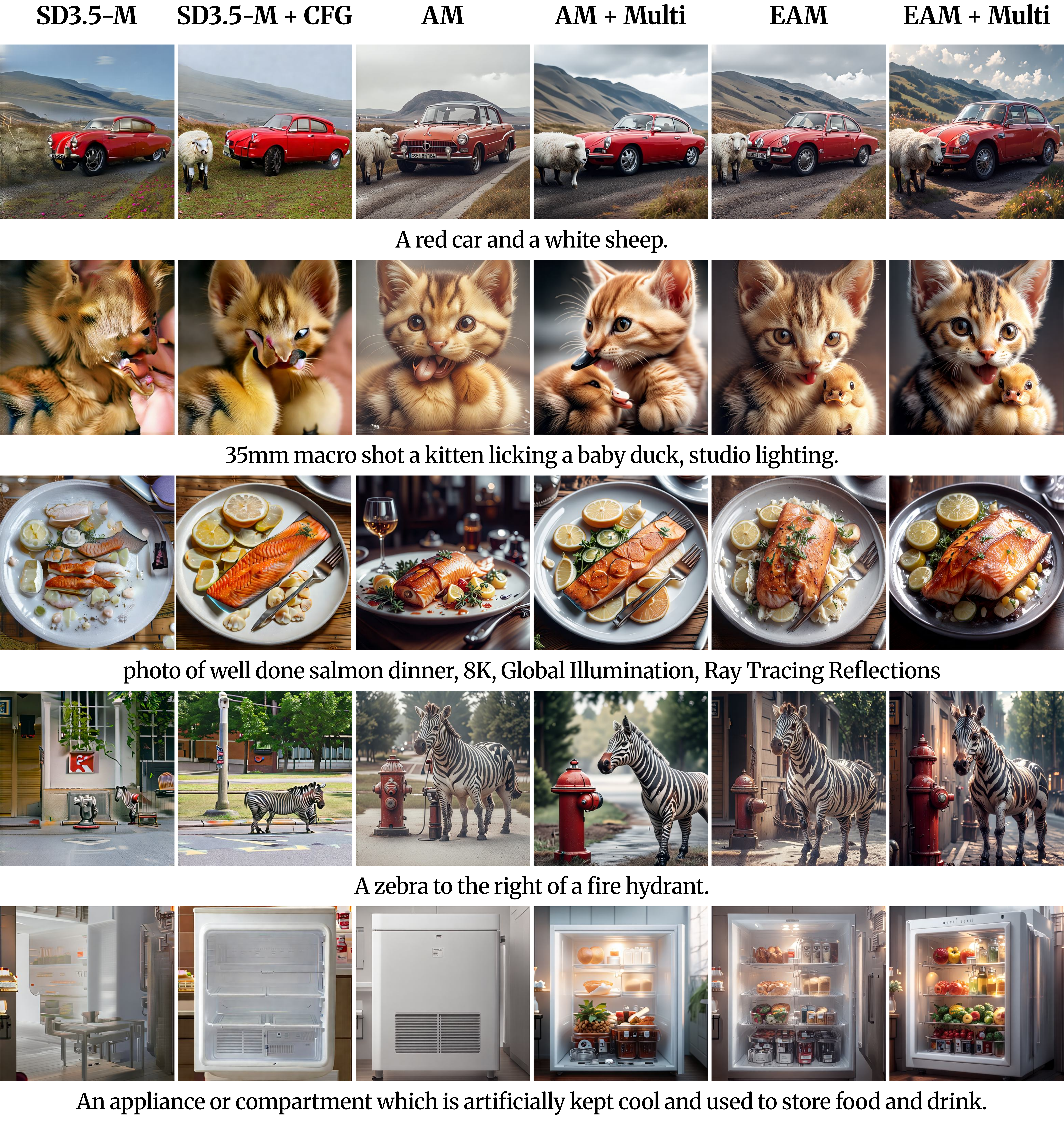}
    \vspace{-.4cm}
    \caption{
        \textbf{Qualitative comparison}. See \cref{appendix:examples} for more examples.
    }
    \label{fig:qualitative_results}
    % \vspace{-1.2cm}
\end{figure}
\subsection{Training Algorithm}
\label{sec:method:alg}
\textbf{Score estimation.}
The terminal cost correction in~\cref{eq:ours_g} requires the pretrained terminal score $\nabla\log p^{\mathrm{pt}}_1(x)$. 
We estimate it using Tweedie's formula~\cite{efron2011tweedie} along the pretrained linear path~\eqref{eq:flow_gen_dynamic}:
\begin{align}
    \nabla\log p^{\mathrm{pt}}_{\tilde t}(x)
    =
    \frac{\tilde t\,v^{\mathrm{pt}}(x,\tilde t)-x}{1-\tilde t},
    \qquad \tilde t\approx 1.
    \label{eq:score_est}
\end{align}
Ideally, this estimate uses a perturbation level $\tilde t$ close to $1$ to approximate $\nabla\log p^{\mathrm{pt}}_1(x)$. 
In practice, we reuse the intermediate state $(X_t,t)$ constructed for the matching loss (via noising kernel $q_t$~\eqref{eq:flow_kernel}) and plug it into~\cref{eq:score_est}, avoiding an additional noising step:
\begin{align}\label{eq:grad_term_cost}
    \nabla g(X_1) \approx -\frac{1}{2C-1}X_1 + \frac{1}{1-t}X_t - \frac{t}{1-t}v^{\text{pt}}(X_t,t)-\beta\nabla r(X_1).
\end{align}
While this introduces bias for $t$ far from $1$, it avoids the unreliable velocity estimates near $t=1$ and yields a lower-variance target. Ablation results comparing this estimator~\eqref{eq:score_est} with the exact Tweedie estimator and an ELBO-based estimator~\cite{choi2026rethinking} are provided in~\cref{appendix:ablation}.

% \jaemoo{Write a full form of $\nabla g(x)$}

% \textbf{Estimation of score.} We estimate $\mathbf\nabla \log p^{\text{pt}}_1(x)$ in $\nabla g(x)$~\eqref{eq:ours_g} via Tweedie's formula~\cite{efron2011tweedie}:
% \begin{align}
%     \nabla\log p^{\text{pt}}_t(x)= \frac{tv^{\text{pt}}(x,t)-x}{1-t},
%     \label{eq:score_est}
% \end{align}
% where timestep $t$ plays as a perturbation level, which controls the trade-off between approximation bias and variance of score estimation~\cite{song2019ncsn}. In practice, we simply use the intermediate state $X_t$~\eqref{eq:alg:xt}.

\textbf{Control parameterization.} A direct parameterization of $u$ is straightforward but starts fine-tuning from a randomly initialized control. To instead initialize from the pretrained model, we rewrite the pretrained generative dynamic~\eqref{eq:flow_gen_dynamic} as a controlled dynamic around the redesigned base drift~\eqref{eq:reformed_base_drift}:
\begin{align}
\scalebox{0.9}{$\displaystyle
    \rd X_t=D(t)X_t \rd t+ \sigma(t)\left(\frac{1}{\sigma(t)}\left(-D(t)X_t-\frac{1}{t}X_t+2v^{\text{pt}}(X_t,t)\right)\right)\rd t+\sigma(t)\rd W_t, \,\,\, X_0\sim \mathcal{N}(0,I).
    \label{eq:ours_sde_reparam}
$}
\end{align}
This decomposition suggests parameterizing the control through a trainable velocity model:
\begin{align}
    u(x,t)
    =
    \frac{1}{\sigma(t)}
    \left(
    -D(t)x-\frac{1}{t}x+2v^{\mathrm{ft}}(x,t)
    \right),
    \label{eq:control_param}
\end{align}
where $v^{\mathrm{ft}}$ is initialized from the pretrained model $v^{\mathrm{pt}}$. 
When $v^{\mathrm{ft}}=v^{\mathrm{pt}}$, the controlled dynamic~\eqref{eq:controlled-sde} exactly recovers the pretrained generative dynamic~\eqref{eq:flow_gen_dynamic}. 
Thus, this parameterization lets fine-tuning start from the pretrained model while learning the control induced by the redesigned base drift.

\textbf{Training objective.}
% \textbf{Algorithm \& practical implementation}\\
% 1. Algorithm explanation: explain each term intuitively\\
% 2. Practical settings: estimation of $\nabla \log p^{\text{pt}}(X_1)$ via approximated smoothed score \& reward coefficient
% % 각 term의 의미 + exp ablation
% \jaemoo{Don't write a full form. Our loss is \cref{eq:am-loss-closed-adj-kernel}, and each term $D(x)$, $u(x,t)$, and $\nabla g(x)$ is defined in \cref{eq:reformed_base_drift}, \cref{eq:control_param}, and Eq. ??. So, now we have a full picture}
Our final loss objective is given by~\cref{eq:am-loss-closed-adj-kernel}, with $p^{\text{base}}_{t|1}$, $D(t)$, $\nabla g(x)$, and $u(x,t)$ specified in~\cref{eq:condition_kernel_match}, \cref{eq:reformed_base_drift}, \cref{eq:grad_term_cost}, and \cref{eq:control_param}, respectively.
% Combining all, our final loss objective is
% \begin{equation}
% \scalebox{1.0}{$\displaystyle
% \begin{gathered}
%     \mathcal{L}_{\mathrm{EAM}}
%     :=
%     \int_0^1
%     \frac{1}{\sigma(t)^2}
%     \left\|
%     v^{\text{ft}}(X_t,t)
%     + \frac{1 - 2Ct}{2Ct^2 - 2t + 1} X_t
%     + \frac{(2C-1)(1-t)}{2Ct^2 - 2t + 1} \nabla g(X_1)
%     \right\|^2 \, \rd t, \\
%     X_t\sim p^{\bar u},
%     \qquad \bar u = \operatorname{stopgrad}(u),
% \end{gathered}
% $}
% \label{eq:training_obj_ours}
% \end{equation}
% where $\nabla g(x)$ is defined as \cref{eq:ours_g}, and $\nabla \log p_t^{\text{pt}}(x)$ is calculated as \cref{eq:score_est}. 

\textbf{Practical implementation.}
For numerical stability, we apply timestep-dependent reward scaling, using $w(t)=(1-t)^{0.9}/t^{1.5}$ for the reward term and its inverse for loss weighting.

\textbf{Summary.} \cref{tab:am-vs-ours} summarizes the difference between AM and our EAM. EAM effectively removes main computational bottlenecks in AM, \emph{i.e.}, adjoint simulation and stochastic trajectory simulation via redesigning the base dynamic. As shown in \cref{alg:ours}, we simulate $X_1$ with efficient ODE solver~\eqref{eq:alg:sampling} and 
% obtain intermediate state $X_t$ via perturbation kernel of original diffusion model~\eqref{eq:alg:xt}. 
construct the intermediate state $X_t$ by sampling from the original noising kernel $q_t(\cdot| X_1)$
Then, we minimize adjoint matching loss~\eqref{eq:am-loss-closed-adj-kernel}, which does not need adjoint ODE simulation.

% \begin{equation}
% \scalebox{0.93}{$\displaystyle
% \begin{gathered}
%     \mathcal{L}_{\mathrm{SAM}}
%     :=
%     \frac{1}{2}
%     \int_0^1
%     \Bigl\|
%     2v^{\text{ft}}(X_t,t)
%     - B_1(t)(X_1 - \epsilon)
%     - B_2(t)\,v^{\text{pt}}(X_t,t)
%     + B_3(t)\,\epsilon
%     - B_3(t)\,\nabla r(X_1)
%     \Bigr\|^2 \, \rd t, \\
%     X_t\sim p^{\bar u},
%     \qquad \bar u = \operatorname{stopgrad}(u), \\
%     \text{where}\quad
%     B_1(t) = \frac{2(1-t)(1-2Ct)}{2Ct^2 - 2t + 1},
%     \quad
%     B_2(t) = \frac{2(2C-1)\,t}{2Ct^2 - 2t + 1},
%     \quad
%     B_3(t) = \frac{2(2C-1)(1-t)}{2Ct^2 - 2t + 1}.
% \end{gathered}
% $}
% \end{equation}
% Intuitively, $X_1-\epsilon$ term explores the unseen velocity while $v^{\text{pt}}$ acts as an anchor to bind $v^{\text{ft}}$ to pretrained generative dynamic, and $\epsilon$ term 
\section{Experiments}
\label{sec:experiments}
\subsection{Experimental Settings}
% \begin{table*}[t]
%     \centering
%     \caption{\textbf{Comparison across evaluation metrics.} All the evalautions were conducted on $4\times$ NVIDIA A100 GPUs.}
%     \label{tab:metric_comparison}
%     \vspace{0.5em}
%     \setlength{\tabcolsep}{6pt}
%     \renewcommand{\arraystretch}{1}
%     \begin{tabular}{lccccc}
%         \toprule
%         Method & PickScore & CLIPScore & ImageReward & HPSv2.1 & Aesthetics \\
%         \midrule
%         SD3.5       & 20.34 & 0.226 & -0.58 & 0.214 & 5.29 \\
%         SD3.5 + CFG & 21.82 & \textbf{0.275} & 0.65 & 0.265 & 5.41 \\
%         AM          & 23.00 & 0.259 & 0.97 & 0.292 & 6.25 \\
%         AM + Triple & 22.63 & 0.256 &  0.84 & 0.305 & 6.04 \\
%         Ours        & \textbf{23.16} & 0.252 &  0.95 & 0.297 & 6.27 \\
%         Ours + Triple & 22.93 & 0.261 & \textbf{0.98} & \textbf{0.306} & \textbf{6.56} \\
%         \bottomrule
%     \end{tabular}
% \end{table*}
\begin{table*}[t]
    \centering
    \caption{\textbf{Quantitative evaluation across metrics.}
    EAM achieves comparable or better alignment quality than AM across most metrics, both when optimizing PickScore alone and when optimizing the combined reward of PickScore, HPSv2.1, and Aesthetics.}
    \label{tab:metric_comparison}
    \vspace{0.5em}
    \setlength{\tabcolsep}{6pt}
    \renewcommand{\arraystretch}{1}
    \begin{tabular}{lccccc}
        \toprule
        Method & PickScore & CLIPScore & ImageReward & HPSv2.1 & Aesthetics \\
        \midrule
        SD3.5-M~\cite{esser2024scaling}         & 20.34 & 0.226 & -0.58 & 0.214 & 5.29 \\
        SD3.5-M + CFG   & 21.82 & \textbf{0.275} & 0.65 & 0.265 & 5.41 \\
        \midrule
        \multicolumn{6}{l}{\textcolor{black}{\textit{PickScore}}} \\
        AM~\cite{domingoenrich2025adjointmatching}            & 23.00 & 0.259 & \underline{0.97} & 0.292 & 6.25 \\
        \textbf{EAM (ours)}          & \textbf{23.16} & 0.252 & 0.95 & 0.297 & \underline{6.27} \\
        \midrule
        \multicolumn{6}{l}{\textcolor{black}{\textit{PickScore + HPSv2.1 + Aesthetics}}} \\
        AM            & \underline{23.06} & 0.253 & 0.83 & \underline{0.303} & 6.21 \\
        \textbf{EAM (ours)}          & 22.93 & \underline{0.261} & \textbf{0.98} & \textbf{0.306} & \textbf{6.56} \\
        \bottomrule
    \end{tabular}
\end{table*}
\textbf{Setup.}
We fine-tune Stable Diffusion 3.5-Medium (SD3.5-M)~\cite{esser2024scaling} by training LoRA~\cite{hu2022lora} weights of rank 32 to generate images at $512\times512$ resolution, using Pick-a-Pic~\cite{kirstain2023pickapic} as the training prompt set. We conduct all experiments on 4 NVIDIA A100 GPUs.
We consider two reward settings: a single-reward setting using PickScore~\cite{kirstain2023pickapic}, and a multi-reward setting combining PickScore, HPSv2.1~\cite{wu2023humanpreference}, and Aesthetics~\cite{schuhmann2022aesthetics}. 
All fine-tuned models are trained for one epoch with an effective batch size of 512 using AdamW~\cite{loshchilov2019decoupledweightdecay}, with learning rate $1\times10^{-4}$ and momentum parameters $(\beta_1,\beta_2)=(0.9,0.999)$. 
AM uses 40 NFEs while EAM uses 10 NFEs for the trajectory simulation. 
Unless otherwise stated, we set the reward scale to $\beta=2000$ for both AM and EAM, and use $C=0.51$ for EAM. 
We evaluate on DrawBench~\cite{saharia2022photorealistic}, generating images with 10 NFEs using DPM-Solver++(2M)~\cite{lu2025dpmsolver}.
 
% \textbf{Evaluation metrics.} We evaluate fine-tuned models using multiple complementary metrics to measure the alignment with human preferences. PickScore \cite{kirstain2023pickapic}, ImageReward \cite{xu2023imagereward}, and HPSv2.1 \cite{wu2023humanpreference} estimate human preference for generated images by jointly considering the text prompt and the generated image. Aesthetics \cite{schuhmann2022aesthetics} measures visual appeal using only image embeddings, while CLIPscore \cite{hessel2021clipscore} quantifies image--text compatibility.

\textbf{Evaluation metrics.} We evaluate fine-tuned models using complementary metrics for human preference and prompt alignment. 
PickScore~\cite{kirstain2023pickapic}, ImageReward~\cite{xu2023imagereward}, and HPSv2.1~\cite{wu2023humanpreference} estimate human preference by jointly considering the prompt and generated image. 
Aesthetics~\cite{schuhmann2022aesthetics} measures visual appeal from image embeddings, while CLIPScore~\cite{hessel2021clipscore} measures image--text compatibility.

\subsection{Main Results.} 
% \begin{wrapfigure}{r}{0.5\textwidth}
%     \vspace{-0.4cm}
%             \begin{center}
%             \hspace{-0.5cm}
%         \includegraphics[height=0.15\textheight]{fig/traintime.pdf}
%             \end{center}
%             \vspace{-0.3cm}
%         \caption{\textbf{PickScore evaluation by GPU hours}. This is temp figure.}
%         \label{fig:attn_nmi}
%     \vspace{-0.4cm}
% \end{wrapfigure}

\begin{wrapfigure}{r}{0.45\textwidth}
    \vspace{-0.6cm}
    \begin{center}
        \includegraphics[width=\linewidth]{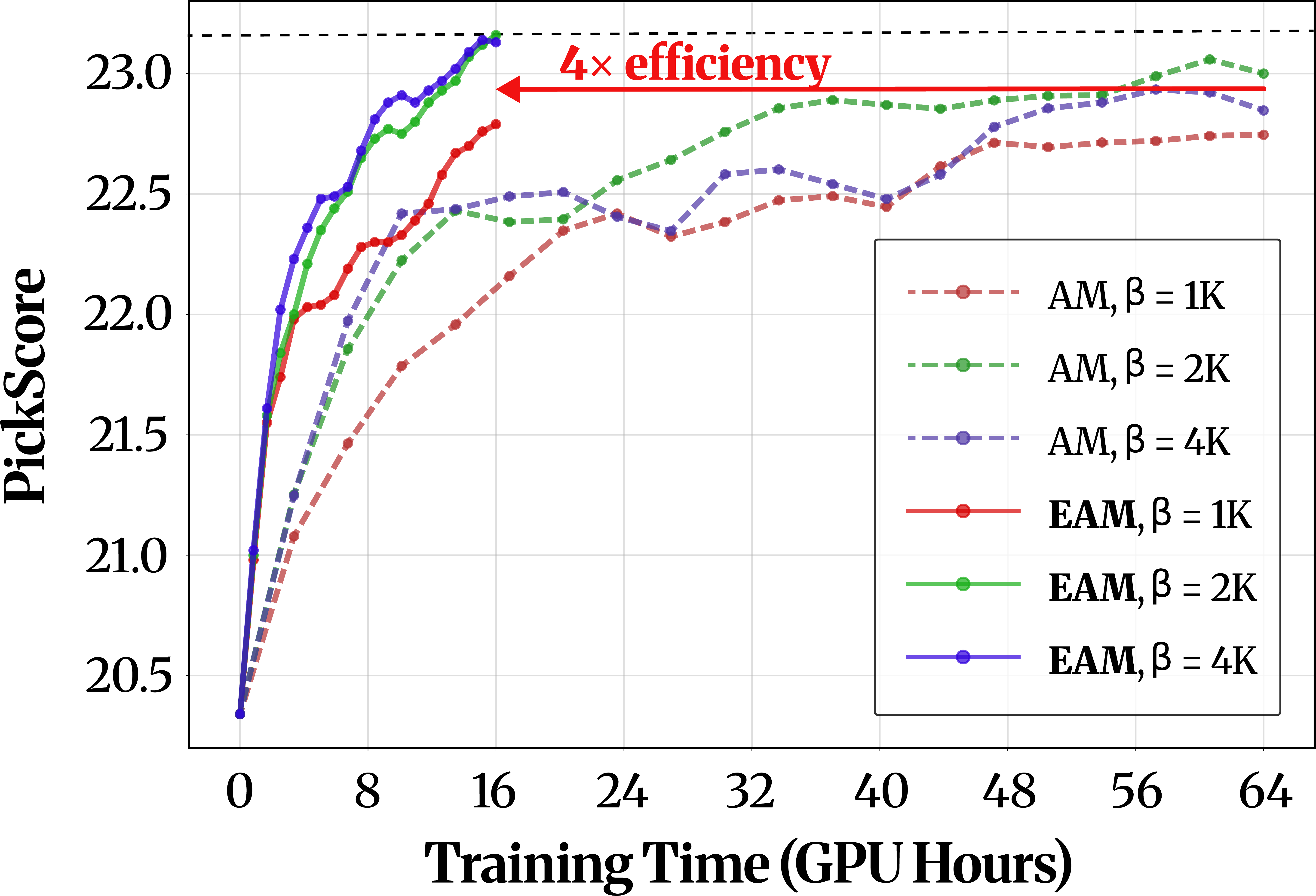}
    \end{center}
    \vspace{-0.3cm}
    \caption{\textbf{PickScore on DrawBench by training time (GPU hours).}
    EAM converges significantly faster than AM (up to 4×).}
    \label{fig:traintime}
    \vspace{-.5cm}
\end{wrapfigure}

\textbf{Quantitative results.}
%\jaemoo{Did you define CFG? Is SD3.5 correct notation?}
As shown in~\cref{tab:metric_comparison}, EAM consistently matches or outperforms AM across most metrics in both the single-reward and multi-reward settings. 
Both EAM and AM substantially improve over the pretrained SD3.5-M baseline, while SD3.5-M with Classifier-Free-Guidance (CFG) ~\cite{ho2022classifierfree} attains the highest CLIPScore. 
Applying CFG to AM and EAM further improves most metrics except Aesthetics, as shown in \cref{tab:appendix_cfg} of \cref{appendix:experiments}. However, it doubles the NFE by requiring both conditional and unconditional velocity evaluations at each sampling step during image generation.
Results on additional backbones and resolutions are provided in~\cref{appendix:sana}.
% shows that our EAM surpasses the baselines across four metrics, except for the CLIPScore. We also report the metrics when Classifier-Free Guidance (CFG) is applied, using guidance scale $\omega=2$ at inference time and following the formula $(1+\omega)v(x_t\mid c)+\omega \, v(x_t)$ from \cite{ho2022classifierfree}. 

\begin{table}[t]
\centering
\caption{\textbf{Source of EAM's efficiency gain} (SD3.5-M).
(a) Matched-NFE comparison.
(b) Per-iteration time and peak memory of the trajectory and adjoint phases only; costs shared by both methods (\emph{e.g.}, reward gradient, loss backpropagation) are excluded.}
\label{tab:efficiency}
\setlength{\tabcolsep}{4pt}
\begin{minipage}[t]{0.48\linewidth}
\centering
\textbf{(a)} Matched NFE\\[0.3em]
\begin{tabular*}{\linewidth}{@{\extracolsep{\fill}}lccc@{}}
\toprule
Method & NFE & GPU hrs $\downarrow$ & PickScore $\uparrow$ \\
\midrule
AM (SDE)  & 40 & 64 & 23.00 \\
AM (SDE)  & 10 & 80 & 23.03 \\
EAM (ODE) & 10 & \textbf{16} & \textbf{23.16} \\
\bottomrule
\end{tabular*}
\end{minipage}\hfill
\begin{minipage}[t]{0.48\linewidth}
\centering
\textbf{(b)} Per-iteration cost\\[0.3em]
\begin{tabular*}{\linewidth}{@{\extracolsep{\fill}}lcccc@{}}
\toprule
& \multicolumn{2}{c}{Time (s) $\downarrow$} & \multicolumn{2}{c}{Mem. (GB) $\downarrow$} \\
Phase & AM & EAM & AM & EAM \\
\midrule
Trajectory & 16.7 & \textbf{4.5}  & 27.4 & \textbf{27.2} \\
Adjoint    & 34.7 & \textbf{0.4} & 73.6 & \textbf{1.6} \\
\bottomrule
\end{tabular*}
\end{minipage}
\end{table}
\textbf{Training efficiency.}
\cref{fig:traintime} shows the trade-off between training time and performance. EAM converges substantially faster than AM, reducing training time by up to 4× while achieving comparable or better performance. This improvement comes from two simplifications: replacing 40-step SDE trajectory simulation with 10-step ODE endpoint sampling, and computing the adjoint matching target in closed-form without backward simulation or costly JVP evaluations.
First, AM cannot use an ODE sampler, as its adjoint must be simulated along a memoryless stochastic trajectory~\eqref{eq:memoryless_cond}, and reducing its SDE steps instead slows convergence by degrading both the samples and the adjoint states simulated along them (\cref{tab:efficiency}a).
In contrast, EAM only requires the endpoint $X_1$, which a few-step ODE solver generates without such degradation, reducing trajectory construction time by 73\% (\cref{tab:efficiency}b).
Second, the closed-form adjoint removes the backward simulation entirely, reducing the adjoint computation time by 99\% and its peak memory from 73.6\,GB to 1.6\,GB.

\textbf{Qualitative results.}
\Cref{fig:qualitative_results} compares images generated by SD3.5-M, AM, and EAM under the single-reward and multi-reward settings. 
Fine-tuning with PickScore improves image completeness for both AM and EAM, reducing visibly distorted or broken samples. 
When HPSv2.1 and Aesthetics are additionally used, both methods produce more polished images with improved prompt fidelity.

% Compared with AM, EAM tends to generate more detailed and compositionally richer images in both reward settings. 
% \jaemoo{Need to discuss this part.}
% We hypothesize that this difference stems from the trajectory construction: EAM trains on endpoint-conditioned intermediate states sampled through the noising kernel in~\cref{eq:condition_kernel_match}, rather than on the exact simulated generative trajectories used by AM. 
% While the latter may provide a more faithful trajectory-specific reward propagation, the former can expose the model to a broader set of intermediate states, potentially improving generalization. See \cref{sec:ablation} for more explanation.

Compared with AM, EAM tends to better preserve fine-grained details and complex compositions in both reward settings. 
We attribute this to two consistency properties of our training procedure. 
First, EAM constructs each intermediate state $X_t$ by sampling from the original noising kernel~\eqref{eq:flow_kernel} conditioned on the generated endpoint $X_1$. 
Thus, the training pair $(X_1,X_t)$ preserves the pretrained diffusion coupling between clean samples $X_1$ and noisy intermediate states $X_t$, aligning with the core perspective of DiffusionNFT~\cite{zheng2026diffusionnft}.
Second, the endpoint distribution used during training matches the one used at evaluation, since both are obtained via ODE simulation with 10 NFEs.
This reduces the mismatch between training and inference trajectories in EAM. 
As a result, EAM improves reward alignment while better retaining detailed structures, whereas AM often improves overall image quality but can smooth out fine details.
% sampling이랑 same scheme (ODE)
% briefly explain the effect of pickscore, triple. explain visual difference between AM and ours. Similar metric scores but shows different style in actual images.

\subsection{Ablation Study}
\label{sec:ablation}

% \textbf{Reward scale.}
% As shown in~\cref{fig:traintime}, the reward scale $\beta$ must be sufficiently large to provide an effective fine-tuning signal, and the appropriate value may depend on the base model, learning rate, reward model, and other training details. 
% In our setting, $\beta=2K$ gives the best overall performance for both AM and EAM. 
% Using a larger value, $\beta=4K$, accelerates early convergence, but its final performance becomes comparable to that of $\beta=2K$.

\textbf{Reward scale.}
As shown in~\cref{fig:traintime}, the reward scale $\beta$ affects both convergence speed and final performance, and its optimal value may depend on the training setup. 
In our setting, $\beta=2000$ gives the best overall performance for both AM and EAM. 
Using a larger value, $\beta=4000$, accelerates early convergence, but its final performance becomes comparable to that of $\beta=2000$.
% In our experiments, $\beta=2K$ performs best for both AM and EAM, while $\beta=4K$ improves early convergence but reaches similar final performance.

\begin{wraptable}{r}{0.38\columnwidth}
\vspace{-.67cm}
\centering
\small
\caption{PickScore for different $C$.}
\label{tab:c_ablation}
\vspace{.2em}
\begin{tabular}{lccc}
\toprule
$C$ & 0.501 & 0.51 & 1.0 \\
\midrule
PickScore & 22.53 & 23.16 & 21.97 \\
\bottomrule
\end{tabular}
\end{wraptable}
% path kl 표현
\textbf{Role of the constant $C$ in balancing loss components.} 
To analyze the effect of $C$ during training, we expand our loss objective in~\cref{eq:am-loss-closed-adj-kernel} and write the matching target for the trainable velocity $v^{\mathrm{ft}}$, omitting the reward term:
\vspace{0.5em}
\begin{align}
    \frac{(1-t)(1-2Ct)}{2Ct^2-2t+1}(X_1-\epsilon)+\frac{t(2C-1)}{2Ct^2-2t+1}v^{\text{pt}}(X_t,t)-\frac{(1-t)(2C-1)}{2Ct^2-2t+1}\epsilon.
\end{align}
% The constant $C$ controls the balance between $X_1-\epsilon$ and the pretrained velocity $v^{\mathrm{pt}}(X_t,t)$: 
% The constant $C$ controls the relative weights of the three non-reward terms: $X_1-\epsilon$, the pretrained velocity $v^{\mathrm{pt}}(X_t,t)$, and the noise term $\epsilon$. 
% Larger $C$ increases the contribution of the pretrained velocity, while smaller $C$ emphasizes $X_1-\epsilon$. 
% The pretrained velocity term keeps $v^{\mathrm{ft}}$ close to the pretrained model, whereas $X_1-\epsilon$ introduces additional exploration at $X_t$.
% In our experiments, $C=0.51$ performs the best; larger values, such as $C\geq 1$, make the $\epsilon$ term overly dominant and lead to unstable optimization.

The constant $C$ balances the three non-reward terms: $X_1-\epsilon$, the pretrained velocity $v^{\mathrm{pt}}(X_t,t)$, and the noise term $\epsilon$. 
Larger $C$ increases the weight on the pretrained velocity, while smaller $C$ emphasizes $X_1-\epsilon$. 
The pretrained velocity keeps $v^{\mathrm{ft}}$ close to the pretrained model, whereas $X_1-\epsilon$ introduces additional exploration at $X_t$. 
In our experiments, $C=0.51$ performs best; larger values, such as $C\geq 1$, make the $\epsilon$ term dominant and destabilize optimization. 
Further ablations on the terminal score estimator, the terminal cost correction, sampling NFE, and LoRA rank are in~\cref{appendix:ablation}.
% \jwchoi{comment}

\section{Related Works}
\label{sec:related_works}
% rel works 대비 우리 contributions
% zero & first -> zero 단점, zero도 추가
% first는 efficient한데 단점, AM 이후 논문들
% AM 되게 좋아, 근데 우리는 더 좋아
% AS와 차이점, base drift 차이 

%Existing reward fine-tuning methods can be grouped into two streams. \emph{Zero-order} methods~\cite{liu2025flowgrpo,zheng2026diffusionnft,choi2026rethinking} treat the reward as a black box and rely solely on its scalar values, adopting policy-gradient-style estimators inherited from reinforcement learning for large language models. \emph{First-order} methods~\cite{clark2024directly,xu2023imagereward,prabhudesai2024aligning,domingoenrich2025adjointmatching}, in contrast, exploit the gradient of the reward with respect to its input, which is readily accessible whenever the reward is parameterized by a differentiable network~\cite{hessel2021clipscore,wu2023humanpreference,schuhmann2022aesthetics,xu2023imagereward,kirstain2023pickapic}, as is typical for learned preference models. When such a gradient is available, first-order methods are substantially more sample-efficient than their zero-order counterparts~\cite{clark2024directly,prabhudesai2024aligning}: each generation trajectory is directly supervised by a full gradient that points toward higher reward, carrying a much richer learning signal than a single scalar return. 
Reward-based fine-tuning methods for diffusion and flow models can be broadly grouped into two streams. \emph{Reward-value-based} methods \cite{black2024training, fan2023dpok, fan2023optimizingddpm, zhao2025scoreaction, liu2025flowgrpo, zheng2026diffusionnft, choi2026rethinking, choi2026scaling} treat the reward as a black box and rely solely on its scalar values, adopting policy-gradient-style estimators inherited from reinforcement learning for large language models. A prominent line of such methods \cite{liu2025flowgrpo, xue2025dancegrpo, he2025tempflowgrpo, wang2025prefgrpo, wang2025grpoguard, ye2025dataregularized, xue2025advantageweight, choi2026rethinking, zheng2026diffusionnft} requires generating 12 to 24 images for each prompt, making the optimization computationally expensive. 
%in order to compute group-relative advantages or determining the positive and negative signals
\emph{Reward-gradient-based} methods~\cite{clark2024directly,xu2023imagereward,prabhudesai2024aligning,guo2025shortft, wu2024deeprewardsupervisionstuning,domingoenrich2025adjointmatching}, in contrast, utilize useful information from the gradient of reward functions. These methods can obtain an update from a single generated sample per prompt, enabling more sample-efficient training. Adjoint Matching (AM) \cite{domingoenrich2025adjointmatching} provides a theoretically grounded SOC formulation for targeting the exact reward-tilted distribution, and is the first to identify the necessity of a memoryless noise schedule in the base dynamic. Subsequent works have extended this framework: ASBS \cite{liu2025adjointschrodingerbridgesampler} generalizes the memoryless condition of the base dynamic, and TR-SOCM \cite{blessing2025trustregion} enables a more stable optimization framework. While promising, training AM-based methods is compute-intensive due to the stochastic simulation for generating images and backward simulation for adjoint states. Our method simultaneously addresses both issues by adequately selecting the linear base drift, while retaining the theoretical guarantees of AM.

Adjoint Sampling \cite{havens2025adjointsampling} is similar in spirit to our work, but it targets a different task and solves it under a distinct setting and formulation. Its goal is to sample from a Boltzmann distribution starting from a Dirac prior, which is achieved by adopting a zero base drift. In contrast, in the standard fine-tuning setup considered in \cref{eq:reward_soc_setting}, the base drift includes the pretrained velocity. Therefore, simplifying the formulation is not immediate and requires satisfying additional structural conditions. Our method addresses this by redesigning the base dynamic so that its drift is linear, the dynamic is memoryless, and its endpoint-conditioned base kernel $p^\text{base}_{t\mid 1}$ matches the original noising kernel $q_{t\mid 1}$. 

\section{Conclusion}
\label{sec:conclusion}
We introduce Efficient Adjoint Matching (EAM), an efficient reward fine-tuning method for diffusion models.
EAM is based on the observation that the two main bottlenecks of AM, forward trajectory simulation and backward adjoint simulation, stem from the choice of the base drift.
By redesigning the base drift to be linear and memoryless, EAM replaces full trajectory simulation with endpoint-conditioned noising and computes the adjoint state in closed-form.
With a rederived terminal cost, the redesigned SOC problem still targets the desired reward-tilted distribution.
Experiments show that EAM matches or improves AM with up to 4× faster convergence.

\textbf{Limitations.}
Exploring practical optimization strategies, such as buffer replay, old-policy trajectory generation, and adaptive hyperparameter schedules, is an important direction for future work.

\section*{Acknowledgments}
We thank Guan-Horng Liu for his assistance, insightful discussions and comments on the manuscript. This research was supported by Youlchon Foundation, the National Research Foundation of Korea (NRF) grants (RS-2021-NR05515, RS-2023-0022663, RS-2024-00336576, RS-2024-00349646, RS-2025-25402648, RS-2025-25406186, RS-2026-25506577), by the Institute for Information \& Communication Technology Planning \& Evaluation (IITP) grants (RS-2022-II220264, RS-2024-00353131), by the National IT Industry Promotion Agency (NIPA) grant (RS-2026-25621677), by the National Institute for Mathematical Sciences (NIMS) grant (No. S26120000), and by the Korea Agency for Infrastructure Technology Advancement (KAIA) grant (RS-2026-25525550) funded by the Korean government.
This research was also supported by the High-Performance Research AI Computing Infrastructure Support (RS-2026-25505492) from IITP. W. G. and Y. C. are supported by NSF 2206576 and NSF 2450378.

% \begin{ack}
% Use unnumbered first level headings for the acknowledgments. All acknowledgments

% Do {\bf not} include this section in the anonymized submission, only in the final paper. You can use the \texttt{ack} environment provided in the style file to automatically hide this section in the anonymized submission.
% \end{ack}

\medskip
{
    \small
    \bibliographystyle{abbrv}
    \bibliography{main}
}

\clearpage
\appendix
\crefalias{section}{appendix}
\crefalias{subsection}{appendix}
\pagenumbering{roman}
\setcounter{page}{1}
\setcounter{section}{0}

\noindent\textbf{\Large Appendix}
% \section{Impact Satement}
% This work develops computational methods for reward-based fine-tuning of diffusion models.
% Our study is theoretical and computational in nature and uses only publicly available image datasets. 
% It does not involve the collection of personal data or the use of sensitive content. 
% Therefore, we do not identify any direct ethical concerns specific to the proposed method. 
% Although generative modeling in general may have broad societal implications, our framework does not introduce additional application-specific risks that warrant separate highlighting.

\section{Proofs}
\subsection{Proof of Closed-form Adjoint in Case of Linear Drift in \cref{eq:closed_adj}}
\label{appendix:proof:linear_drift_adj}

By definition, the lean adjoint state $a(t;X_t)$ satisfies the ODE in~\cref{eq:adj-ode}:
\begin{align}
    \frac{\rd}{\rd t}\,a(t;X_t) \;=\; -\,\nabla_x b(X_t,t)^{\top}\,a(t;X_t),\qquad a(1;X_1)=\nabla g(X_1).
\end{align}
When the base drift takes the linear form $b(x,t)=D(t)\,x$, its Jacobian is $\nabla_x b(X_t,t)=D(t)\,I$, so the ODE reduces to a scalar linear ODE in $a(t;X_t)$:
\begin{align}
    \frac{\rd}{\rd t}\,a(t;X_t) \;=\; -\,D(t)\,a(t;X_t).
\end{align}
Integrating backward from $t=1$ with terminal state $a(1;X_1)$ yields
\begin{align}
    a(t;X_t) \;=\; \exp\!\left(\int_t^1 D(\tau)\,\rd\tau\right) a(1;X_1),
\end{align}
which is~\cref{eq:closed_adj}. The prefactor depends only on $t$, so $a(t;X_t)$ is determined by the endpoint $X_1$ alone via $\nabla g(X_1)$, with no JVP through the velocity network. \qed

\subsection{Proof of $(X_t,X_1)$ Factorization in \cref{eq:reciprocal}}
\label{appendix:proof:linear_drift_recip}

We show that the joint distribution $p^{u}(X_t,X_1)$ admits the factorization
\begin{align}
    p^{u}(X_t,X_1) \;=\; p^{\mathrm{base}}(X_t\mid X_1)\,p^{u}_1(X_1),
    \label{eq:appendix:rp_target}
\end{align}
under the optimal control $u^\star$.

\textbf{Step 1: reciprocal projection.}
Marginalizing the joint $(X_t,X_0,X_1)$ over $X_0$ gives
\begin{align}
    p^{u}(X_t,X_1) \;=\; \int p^{u}(X_t\mid X_0,X_1)\,p^{u}(X_0,X_1)\,\rd X_0.
\end{align}
Reciprocal projection~\cite{shi2023dsbm, havens2025adjointsampling} replaces the controlled bridge by the base bridge,
\begin{align}
    p^{u}(X_t\mid X_0,X_1) \;=\; p^{\mathrm{base}}(X_t\mid X_0,X_1),
\end{align}
which holds at optimality. Then, the joint distribution becomes
\begin{align}
    p^{u}(X_t,X_1) \;=\; \int p^{\mathrm{base}}(X_t\mid X_0,X_1)\,p^{u}(X_0, X_1)\,\rd X_0.
    \label{eq:appendix:rp_intermediate}
\end{align}

\textbf{Step 2: memorylessness at optimality.}
When the base dynamic is memoryless~\eqref{eq:memoryless_cond}, controlled dynamic with optimal control is also memoryless~\cite{shin2026asbm}, \emph{i.e.},
\begin{align}
    p^{u^\star}(X_0,X_1)\overset{\textnormal{memoryless}}{=}p^{\text{base}}_0(X_0)\,p^{u^\star}_1(X_1).
    \label{eq:appendix:rp_endpoints}
\end{align}
As a result, under the SOC optimality, the joint distribution factors as
\begin{align}
    p^{u}(X_t,X_1) \;&=\; \int p^{\mathrm{base}}(X_t\mid X_0,X_1)\,p^{u}(X_0, X_1)\,\rd X_0\\
    \;&=\; \int \frac{p^{\mathrm{base}}(X_t, X_0,X_1)}{p^{\mathrm{base}}_0(X_0)p^{\mathrm{base}}_1(X_1)}\,p^{\mathrm{base}}_0(X_0)p^{u}_1(X_1)\,\rd X_0\\
    \;&=\; p^{\mathrm{base}}(X_t\mid X_1)\,p^{u}(X_1),
\end{align}
which is~\cref{eq:appendix:rp_target}. \qed

\subsection{Proof of \cref{prop:reformed_linear_drift}}
\label{appendix:proof:reformed_linear_drift}

We work with the linear base SDE
\begin{align}
    \rd X_t \;=\; D(t)\,X_t\,\rd t \;+\; \sigma(t)\,\rd W_t,\qquad X_0\sim\mathcal{N}(0,I),\qquad \sigma(t)=\sqrt{\tfrac{2(1-t)}{t}}.
    \label{eq:appendix:linear_sde}
\end{align}
Throughout this proof, we use the abbreviations
\begin{align}
    \Phi_t \;:=\; \exp\!\left(\int_t^1 D(s)\,\rd s\right),\qquad
    I_t \;:=\; \int_0^t \exp\!\left(2\!\int_r^t D(u)\,\rd u\right)\!\frac{1-r}{r}\,\rd r,
\end{align}
and adopt the standard convention $J_t := I_1 - I_t\,\Phi_t^2$.

\paragraph{Step 1: bridge of a linear SDE.}
The solution of~\cref{eq:appendix:linear_sde} is $X_t=\alpha_t X_0 + \beta_t\,\varepsilon$ with $\varepsilon\sim\mathcal{N}(0,I)$, where $\alpha_t=\exp(\int_0^t D)$ and $\beta_t^2=\sigma(t)^2$-driven variance. A direct calculation gives the bridge distribution
\begin{align}
    p(X_t\mid X_0,X_1) \;=\; \mathcal{N}\!\bigl(\bar\alpha_t X_0 + \bar\beta_t X_1,\;\gamma_t^2\, I\bigr),
\end{align}
with bridge coefficients $\bar\alpha_t=\Phi_0 J_t/(\Phi_t I_1)$, $\bar\beta_t=\Phi_t I_t/I_1$, and $\gamma_t^2 = 2\,I_t J_t/I_1$.

\paragraph{Step 2: reciprocal kernel under Gaussian prior.}
Since $X_0\sim\mathcal{N}(0,I)$, marginalizing over $X_0$ yields
\begin{align}
    X_t\mid X_1 \;\sim\; \mathcal{N}\!\bigl(\bar\beta_t X_1,\;(\bar\alpha_t^{2}+\gamma_t^{2})\,I\bigr).
    \label{eq:appendix:reciprocal_kernel}
\end{align}

\paragraph{Step 3: matching the perturbation kernel.}
We require~\cref{eq:appendix:reciprocal_kernel} to coincide with the perturbation kernel $\mathcal{N}(tX_1,(1-t)^2 I)$ of~\cref{eq:flow_kernel}, i.e.,
\begin{align}
    \bar\beta_t = t,\qquad \bar\alpha_t^{2}+\gamma_t^{2} = (1-t)^2.
    \label{eq:appendix:two_constraints}
\end{align}

\paragraph{Step 4: solving the first constraint.}
Substituting $\bar\beta_t=\Phi_t I_t/I_1$, the constraint $\bar\beta_t=t$ becomes $\Phi_t I_t = t\,I_1$. Differentiating both sides in $t$ and using Leibniz's rule with $\Phi_t' = -D(t)\,\Phi_t$ and $I_t' = (1-t)/t + 2D(t)\,I_t$, we obtain
\begin{align}
    I_1 \;=\; \Phi_t\!\left(\frac{1-t}{t} + D(t)\,I_t\right),
\end{align}
which, after substituting $I_1=\Phi_t I_t/t$, gives a closed-form expression for $D(t)$:
\begin{align}
    D(t) \;=\; \frac{I_t-(1-t)}{t\,I_t}.
    \label{eq:appendix:D_in_It}
\end{align}
Plugging~\cref{eq:appendix:D_in_It} back into the ODE for $I_t$ yields
\begin{align}
    \frac{\rd I_t}{\rd t} - \frac{2}{t}\,I_t \;=\; -\frac{1-t}{t}.
\end{align}
Multiplying by the integrating factor $t^{-2}$ gives $\frac{\rd}{\rd t}(t^{-2}I_t) = -(1-t)/t^3$, and integrating produces
\begin{align}
    I_t \;=\; \tfrac{1}{2} - t + C\,t^2,\qquad C\in\mathbb{R}.
\end{align}
The constraint $I_t>0$ on $(0,1)$ requires $C>\tfrac{1}{2}$. Substituting this $I_t$ into~\cref{eq:appendix:D_in_It} yields the announced family
\begin{align}
    D(t) \;=\; \frac{2Ct^2-1}{t\,(2Ct^2-2t+1)},\qquad C>\tfrac{1}{2},
\end{align}
which is~\cref{eq:reformed_base_drift}.

\paragraph{Step 5: the second constraint is automatic.}
Using the explicit forms of $I_t$, $\Phi_t = (2C-1)\,t/(2Ct^2-2t+1)$, and $J_t = I_1 - I_t\,\Phi_t^2$, a direct computation (using $\bar\alpha_t = \Phi_0 J_t/(\Phi_t I_1)$ and $\gamma_t^2 = 2I_t J_t/I_1$) verifies that $\bar\alpha_t^{2}+\gamma_t^{2}=(1-t)^2$ holds identically for every $C>\tfrac{1}{2}$. The second constraint in~\cref{eq:appendix:two_constraints} therefore imposes no additional restriction on $D(t)$.

\paragraph{Step 6: terminal distribution.}
At $t=1$, $\Phi_0$ is the value of $\Phi$ at $t=0$, which evaluates to $\Phi_0=0$ (since the numerator $(2C-1)\,t$ vanishes at $t=0$). Hence $\bar\alpha_1=0$, and $X_1 = \bar\beta_1 X_1 + \gamma_1\,\varepsilon$ marginalizes (using $X_0\sim\mathcal{N}(0,I)$ and $\bar\beta_1=1$) to
\begin{align}
    p^{\mathrm{base}}_1 \;=\; \mathcal{N}\!\bigl(0,\,(2C-1)\,I\bigr),
\end{align}
since $I_1 = C - \tfrac{1}{2}$ and $2I_1 = 2C-1$.

\paragraph{Step 7: memorylessness and uniqueness.}
The joint $(X_0,X_1)$ has $\bar\alpha_1=0$, so $X_1$ depends on $X_0$ only through $\bar\beta_1 X_1$ and the independent noise $\gamma_1\varepsilon$; equivalently, $p^{\mathrm{base}}_{0,1}=p^{\mathrm{base}}_0\,p^{\mathrm{base}}_1$, which is the memoryless condition~\eqref{eq:memoryless_cond}.

For the converse, the derivation of Steps 4--5 shows that any linear drift $D(t)$ with the fixed $\sigma(t)$ that satisfies~\cref{eq:appendix:two_constraints} must take the form of~\cref{eq:reformed_base_drift} for some $C>\tfrac{1}{2}$. The family is therefore unique up to the single scalar $C$. \qed

\textbf{Exact adjoint calculation.}
Plugging the linear base drift in~\cref{eq:reformed_base_drift} into~\cref{eq:closed_adj} yields
\begin{align}
    a(t;X_t)=\frac{(2C-1)\,t}{2Ct^2-2t+1}\,a(1;X_1), \qquad a(1;X_1)=\nabla g(X_1).
    \label{eq:closed_form_adjoint_ours}
\end{align}

\subsection{Proof of \cref{prop:terminal_cost}}
\label{appendix:proof:terminal_cost}

We use the standard SOC reduction for reward-tilted sampling~\cite{domingoenrich2025adjointmatching,havens2025adjointsampling}: under the SOC problem~\eqref{eq:control-objective}--\eqref{eq:controlled-sde} with terminal cost $g$, the optimal control $u^\star$ steers the controlled dynamic so that its terminal distribution is
\begin{align}
    p^{u^\star}_1(x) \;\propto\; p^{\mathrm{base}}_1(x)\,\exp\bigl(-g(x)\bigr).
    \label{eq:appendix:soc_reduction}
\end{align}

\paragraph{Designing $g$ for a target tilt.}
We want the terminal distribution to be the reward-tilted target $p^{u^\star}_1(x)\propto p^{\mathrm{pt}}_1(x)\exp(\beta r(x))$. Equating with~\cref{eq:appendix:soc_reduction},
\begin{align}
    p^{\mathrm{base}}_1(x)\,\exp\bigl(-g(x)\bigr) \;\propto\; p^{\mathrm{pt}}_1(x)\,\exp\bigl(\beta r(x)\bigr),
\end{align}
which gives, up to an additive constant absorbed in normalization,
\begin{align}
    g(x) \;=\; \log p^{\mathrm{base}}_1(x) \;-\; \log p^{\mathrm{pt}}_1(x) \;-\; \beta\,r(x).
    \label{eq:appendix:terminal_cost_general}
\end{align}
Taking the gradient,
\begin{align}
    \nabla g(x) \;=\; \nabla\log p^{\mathrm{base}}_1(x) \;-\; \nabla\log p^{\mathrm{pt}}_1(x) \;-\; \beta\,\nabla r(x).
    \label{eq:appendix:terminal_cost_grad}
\end{align}

\paragraph{Specialization to AM.}
Under the standard SOC-based reward fine-tuning setup of~\cref{eq:reward_soc_setting}, the base dynamic coincides with the pretrained generative dynamic, hence $p^{\mathrm{base}}_1=p^{\mathrm{pt}}_1$. The first two terms in~\cref{eq:appendix:terminal_cost_grad} cancel, and we recover the AM gradient
\begin{align}
    \nabla g(x) \;=\; -\beta\,\nabla r(x).
\end{align}

\paragraph{Specialization to our reformulation.}
By~\cref{prop:reformed_linear_drift}, our reformulated base dynamic has terminal distribution $p^{\mathrm{base}}_1=\mathcal{N}\!\bigl(0,(2C-1)I\bigr)$, so
\begin{align}
    \nabla\log p^{\mathrm{base}}_1(x) \;=\; -\frac{x}{2C-1}.
\end{align}
Substituting into~\cref{eq:appendix:terminal_cost_grad} yields
\begin{align}
    \nabla g(x) \;=\; -\frac{x}{2C-1} \;-\; \nabla\log p^{\mathrm{pt}}_1(x) \;-\; \beta\,\nabla r(x),
\end{align}
which is~\cref{eq:ours_g}. \qed

\subsection{Proof of \cref{thm:path_equiv}}
\label{appendix:proof:path_equiv}

\paragraph{Setup.}
We use the superscript $m\in\{\mathrm{AM},\mathrm{EAM}\}$ to distinguish the two SOC problems, and write $p^{\mathrm{ref},m}$ for the corresponding base path measure:
\begin{itemize}[leftmargin=2em]
    \item AM: $p^{\mathrm{ref},\mathrm{AM}}=p^{\mathrm{pt}}$, with drift $b^{\mathrm{AM}}(x,t)=-\frac{x}{t}+2v^{\mathrm{pt}}(x,t)$ and terminal cost $g^{\mathrm{AM}}=-\beta r$~\eqref{eq:reward_soc_setting};
    \item EAM: $p^{\mathrm{ref},\mathrm{EAM}}=p^{\mathrm{base}}$, with drift $b^{\mathrm{EAM}}(x,t)=D(t)x$~\eqref{eq:reformed_base_drift} and terminal cost $g^{\mathrm{EAM}}$ in~\cref{eq:ours_g}.
\end{itemize}
Both share the diffusion coefficient $\sigma(t)$ and the initial distribution $p_0=\mathcal{N}(0,I)$.
As in~\cref{sec:preliminaries}, we adopt the standard idealization that $v^{\mathrm{pt}}$ is exact. Under this idealization, the pretrained dynamic~\eqref{eq:flow_gen_dynamic} is the time reversal of the Markov noising process with kernel $q_t$, so that $p^{\mathrm{pt}}_{t|1}=q_t$ and $p^{\mathrm{pt}}_t(x)=\int q_t(x\mid x_1)\,p_{\mathrm{data}}(x_1)\,\rd x_1$.
For EAM, $p^{\mathrm{base}}_{t|1}=q_t$ holds by \textbf{Condition 3}~\eqref{eq:condition_kernel_match}.
Thus, in both cases,
\begin{align}
    p^{\mathrm{ref},m}_{t|1}=q_t,
    \qquad
    p^{\mathrm{ref},m}_{0,1}=p_0\,p^{\mathrm{ref},m}_1,
    \label{eq:appendix:path_common}
\end{align}
where the second identity is the memoryless condition~\eqref{eq:memoryless_cond}.

\paragraph{Step 1: optimal endpoint coupling.}
The optimal path measure of the SOC problem~\eqref{eq:control-objective}--\eqref{eq:controlled-sde} is an endpoint tilt of the base path measure~\cite{domingoenrich2025adjointmatching}:
\begin{align}
    p^{u^\star_m}(X)\;\propto\;p^{\mathrm{ref},m}(X)\,e^{-g^m(X_1)}.
    \label{eq:appendix:path_tilt}
\end{align}
Marginalizing onto $(X_0,X_1)$ and using memorylessness~\eqref{eq:appendix:path_common},
\begin{align}
    p^{u^\star_m}_{0,1}(x_0,x_1)
    \;\propto\;
    p_0(x_0)\,p^{\mathrm{ref},m}_1(x_1)\,e^{-g^m(x_1)}
    \;\propto\;
    p_0(x_0)\,p^\star_1(x_1),
\end{align}
where the last step uses $p^{\mathrm{pt}}_1e^{\beta r}\propto p^\star_1$ for AM and~\cref{prop:terminal_cost} for EAM.
Hence, in both cases, $X_0$ and $X_1$ are independent at optimality with $X_1\sim p^\star_1$.

\paragraph{Step 2: optimal single-time marginals coincide.}
By~\cref{eq:appendix:path_tilt}, the optimal path measure shares the base bridge $p^{\mathrm{ref},m}_{t|0,1}$. Combining this with Step 1 and~\cref{eq:appendix:path_common},
\begin{align}
    p^{\star,m}_t(x)
    =\iint p^{\mathrm{ref},m}_{t|0,1}(x\mid x_0,x_1)\,p_0(x_0)\,p^\star_1(x_1)\,\rd x_0\,\rd x_1
    =\int q_t(x\mid x_1)\,p^\star_1(x_1)\,\rd x_1,
\end{align}
where the inner integral over $x_0$ equals $p^{\mathrm{ref},m}_{t|1}=q_t$ because $X_0$ and $X_1$ are independent under the base.
The right-hand side does not depend on $m$, so $p^{\star,\mathrm{AM}}_t=p^{\star,\mathrm{EAM}}_t=:p^\star_t$ for all $t\in(0,1)$.

\paragraph{Step 3: optimal drifts coincide.}
The optimal control is $u^\star_m=\sigma\nabla_x\log h^m$ with $h^m(x,t)=\mathbb{E}_{p^{\mathrm{ref},m}}[e^{-g^m(X_1)}\mid X_t=x]$. By~\cref{eq:appendix:path_tilt} and Bayes' rule, $p^\star_t\propto h^m_t\,p^{\mathrm{ref},m}_t$. Hence the optimal drift $b^m_\star=b^m+\sigma u^\star_m$ satisfies
\begin{align}
    b^m_\star(x,t)
    =\Bigl[b^m(x,t)-\sigma^2(t)\nabla_x\log p^{\mathrm{ref},m}_t(x)\Bigr]
    +\sigma^2(t)\nabla_x\log p^\star_t(x).
    \label{eq:appendix:opt_drift}
\end{align}
By Step 2, the last term is common to both methods. It remains to compare the bracketed term.

\emph{AM.} By Tweedie's formula~\eqref{eq:score_est}, $\sigma^2(t)\nabla_x\log p^{\mathrm{pt}}_t(x)=2v^{\mathrm{pt}}(x,t)-\frac{2x}{t}$, so
\begin{align}
    b^{\mathrm{AM}}(x,t)-\sigma^2(t)\nabla_x\log p^{\mathrm{pt}}_t(x)
    =\Bigl(-\frac{x}{t}+2v^{\mathrm{pt}}(x,t)\Bigr)-\Bigl(2v^{\mathrm{pt}}(x,t)-\frac{2x}{t}\Bigr)
    =\frac{x}{t}.
\end{align}

\emph{EAM.} Since $p^{\mathrm{base}}_{t|1}=q_t$ and $p^{\mathrm{base}}_1=\mathcal{N}(0,(2C-1)I)$, we have $p^{\mathrm{base}}_t=\mathcal{N}(0,N(t)I)$ with $N(t):=t^2(2C-1)+(1-t)^2=2Ct^2-2t+1$. Therefore,
\begin{align}
    b^{\mathrm{EAM}}(x,t)-\sigma^2(t)\nabla_x\log p^{\mathrm{base}}_t(x)
    =\frac{2Ct^2-1}{t\,N(t)}\,x+\frac{2(1-t)}{t\,N(t)}\,x
    =\frac{x}{t}.
\end{align}

In both cases the bracketed term equals $x/t$: the pretrained velocity cancels for AM, and all $C$-dependence cancels for EAM. Substituting into~\cref{eq:appendix:opt_drift},
\begin{align}
    b^{\mathrm{AM}}_\star(x,t)=b^{\mathrm{EAM}}_\star(x,t)=\frac{x}{t}+\sigma^2(t)\nabla_x\log p^\star_t(x).
\end{align}

\paragraph{Step 4: conclusion.}
The two optimal processes have the same drift, the same diffusion coefficient $\sigma(t)$, and the same initial distribution $p_0$. Under standard well-posedness assumptions, uniqueness in law for this SDE implies $p^{u^\star_{\mathrm{AM}}}=p^{u^\star_{\mathrm{EAM}}}$, and hence all multi-time joint distributions coincide. \qed

\section{Experiment Details}
\label{appendix:experiments}
\paragraph{Detailed Setup.}
We fine-tune Stable Diffusion 3.5-Medium (SD3.5-M)~\cite{esser2024scaling} by training LoRA~\cite{hu2022lora} weights of rank 32 to generate images at $512\times512$ resolution, using Pick-a-Pic~\cite{kirstain2023pickapic} as the training prompt set. We conduct all experiments on 4 NVIDIA A100 GPUs.
We consider two reward settings: a single-reward setting using PickScore~\cite{kirstain2023pickapic}, and a multi-reward setting combining PickScore, HPSv2.1~\cite{wu2023humanpreference}, and Aesthetics~\cite{schuhmann2022aesthetics}. 
% For the multi-reward setting reported in \cref{tab:metric_comparison},
% we use a weighted combination of PickScore$=1$, HPSv2.1$=0.5\times26$, Aesthetics$=0.02\times26$ for both AM and EAM,
% and set the reward scaling constant $\beta=1000$ for AM and $\beta=2000$ for EAM.
For multi-reward setting reported in \cref{tab:metric_comparison}, 
we set the reward scaling constant $\beta=1000$, weighted combination of PickScore$=1$, HPSv2.1$=2\times26$, Aesthetics$=0.05\times26$ for AM and $\beta=2000$, weighted combination of PickScore$=1$, HPSv2.1$=0.5\times26$, Aesthetics$=0.02\times26$ for EAM. 
% The multi-reward coefficients were chosen to balance the gradient norms across rewards and demonstrate strong empirical performance.
The multi-reward coefficients and reward scaling constants were chosen to balance the gradient norms across rewards and were individually tuned for AM and EAM to achieve the best empirical performance for each method.
All fine-tuned models are trained for one epoch with an effective batch size of 512 using AdamW~\cite{loshchilov2019decoupledweightdecay}, with learning rate $1\times10^{-4}$ and momentum parameters $(\beta_1,\beta_2)=(0.9,0.999)$. 
AM uses 40 NFEs for stochastic trajectory simulation, while EAM uses 10 NFEs for the ODE solver that samples $X_1$. 
For computing the matching loss in AM, we use four random intermediate states per trajectory, which improves the performance than using a single state. 
Unless otherwise stated, we set the reward scale to $\beta=2000$ for both AM and EAM, and use $C=0.51$ for EAM.  
We evaluate all models on DrawBench~\cite{saharia2022photorealistic}, generating images with 10 NFEs using DPM-Solver++(2M)~\cite{lu2025dpmsolver}.

\paragraph{Classifier-Free Guidance.}
We demonstrate the effect of applying Classifier-Free Guidance (CFG) \cite{ho2022classifierfree} in this section. CFG is known to improve alignment between the generated image and the text prompt, but this comes at the cost of doubled NFEs required to compute the unconditional prediction. As shown in \cref{tab:appendix_cfg}, applying CFG improves all the metrics except Aesthetics. 
\begin{table*}[t]
    \centering
    \caption{\textbf{Ablation over CFG.} We apply the same guidance scale $\omega=2$ at inference time and following the formula $(1+\omega)v(x_t\mid c)-\omega \, v(x_t)$ from \cite{ho2022classifierfree}. Applying CFG improves all the metrics except Aesthetics. }
    \label{tab:appendix_cfg}
    \vspace{0.5em}
    \setlength{\tabcolsep}{6pt}
    \renewcommand{\arraystretch}{1}
    \begin{tabular}{lccccc}
        \toprule
        Method & PickScore & CLIPScore & ImageReward & HPSv2.1 & Aesthetics \\
        \midrule
        SD3.5-M       & 20.34 & 0.226 & -0.58 & 0.214 & 5.29 \\
        SD3.5-M + CFG & 21.82 & \textbf{0.275} & 0.65 & 0.265 & 5.41 \\
        AM          & 23.00 & 0.259 & 0.97 & 0.292 & \underline{6.25} \\
        AM + CFG & 23.05 & \textbf{0.275} & \underline{1.15} & \underline{0.297} & 5.90 \\
        Ours        & \underline{23.16} & 0.252 &  0.95 & \underline{0.297} & \textbf{6.27} \\
        Ours + CFG & \textbf{23.25} & \underline{0.273} & \textbf{1.19} & \textbf{0.309} & 5.99 \\
        \bottomrule
    \end{tabular}
\end{table*}

\section{Additional Experiments}
\label{appendix:additional_exp}

\subsection{Additional Backbones and Resolutions}
\label{appendix:sana}
To verify that EAM generalizes beyond SD3.5-M, we further fine-tune SANA 1.0 (1.6B, $512\times512$)~\cite{xie2024sana} and SANA 1.5 (1.6B, $1024\times1024$)~\cite{xie2025sana15}, which adopt a Linear DiT architecture instead of the MMDiT of SD3.5-M.
We use PickScore as the reward and keep all hyperparameters identical to the SD3.5-M setting in~\cref{appendix:experiments}, including $C=0.51$, without re-tuning.
As shown in~\cref{tab:appendix_sana}, EAM matches or outperforms AM on most metrics for both backbones while requiring 4$\times$ fewer GPU hours, consistent with the SD3.5-M results.

\begin{table}[h]
\centering
\caption{\textbf{Results on SANA 1.0 and SANA 1.5.} All setting is identical to the SD3.5-M setting.}
\label{tab:appendix_sana}
\setlength{\tabcolsep}{4pt}
\begin{tabular}{lcccccc}
\toprule
Method & GPU hrs $\downarrow$ & PickScore & CLIPScore & ImageReward & HPSv2.1 & Aesthetics \\
\midrule
\multicolumn{7}{l}{\emph{SANA 1.0 (512px)}} \\
SANA 1.0       & --  & 21.33 & 0.242 & 0.01 & 0.235 & 5.59 \\
SANA 1.0 + CFG & --  & 22.55 & \textbf{0.275} & \textbf{1.05} & 0.296 & 5.70 \\
AM             & \underline{36}  & \underline{23.24} & 0.265 & 1.03 & \underline{0.300} & \underline{6.02} \\
\textbf{EAM (ours)} & \textbf{9} & \textbf{23.30} & \underline{0.266} & \underline{1.04} & \textbf{0.302} & \textbf{6.15} \\
\midrule
\multicolumn{7}{l}{\emph{SANA 1.5 (1024px)}} \\
SANA 1.5       & --  & 21.49 & 0.249 & 0.16 & 0.244 & 5.67 \\
SANA 1.5 + CFG & --  & 22.59 & \textbf{0.277} & 1.00 & 0.291 & 5.75 \\
AM             & \underline{120} & \underline{23.36} & \underline{0.273} & \textbf{1.08} & \underline{0.300} & \underline{6.14} \\
\textbf{EAM (ours)} & \textbf{30} & \textbf{23.43} & 0.267 & \underline{1.05} & \textbf{0.305} & \textbf{6.25} \\
\bottomrule
\end{tabular}
\end{table}

\subsection{Ablation Studies}
\label{appendix:ablation}

\paragraph{Terminal score estimator.}
On SD3.5-M, we compare our estimator in~\cref{eq:grad_term_cost} with two alternatives for $\nabla\log p^{\mathrm{pt}}_1(X_1)$.
The \emph{exact Tweedie} estimator applies~\cref{eq:score_est} to a separately noised state $\tilde X_{\tilde t}=\tilde tX_1+(1-\tilde t)\epsilon$ with $\tilde t\approx 1$.
The \emph{ELBO} estimator~\cite{choi2026rethinking} approximates the log-likelihood by the ELBO and differentiates it with respect to $X_1$:
\begin{align}
    \nabla\log p^{\mathrm{pt}}_1(X_1)
    \approx
    -\nabla_{X_1}\mathbb{E}_{\tilde t,\epsilon}\!\left[\bigl\|v^{\mathrm{pt}}(\tilde X_{\tilde t},\tilde t)-(X_1-\epsilon)\bigr\|^2\right],
    \qquad
    \tilde X_{\tilde t}=\tilde tX_1+(1-\tilde t)\epsilon.
\end{align}
As shown in~\cref{tab:appendix_score}, both alternatives improve over the pretrained model, but our estimator performs best. 
Moreover, despite their improved PickScore, both alternatives often produce visibly corrupted images in practice, whereas ours does not.
The exact Tweedie estimator relies on the velocity near $\tilde t\approx1$, where the pretrained velocity is less reliable and its error is amplified by $1/(1-\tilde t)$.
The ELBO estimator affords only a few Monte Carlo samples per example, yielding a high-variance regression target that destabilizes training~\cite{kim2026equivariant}, and requires differentiating through $v^{\mathrm{pt}}$.
Our estimator avoids both issues by reusing $(X_t,t)$, yielding a lower-variance target at the cost of some bias.

\begin{table}[h]
\centering
\caption{\textbf{Terminal score estimator} (SD3.5-M).}
\label{tab:appendix_score}
\begin{tabular}{lcccc}
\toprule
& SD3.5-M & Exact Tweedie & ELBO & Ours \\
\midrule
PickScore & 20.34 & 21.78 & 21.54 & \textbf{23.16} \\
\bottomrule
\end{tabular}
\end{table}

\paragraph{Terminal cost correction.}
On SANA 1.0, we examine the role and robustness of the correction term in~\cref{eq:ours_g}.
First, we remove the correction and keep only the reward term, \emph{i.e.}, $\nabla g=-\beta\nabla r$. By~\cref{prop:terminal_cost}, the target then collapses to a reward-tilted Gaussian $p^{\mathrm{base}}_1e^{\beta r}$, and performance degrades substantially (\cref{tab:appendix_correction}a).
Second, we perturb the estimated score $\hat s$ as $\hat s+\alpha\,\mathrm{RMS}(\hat s)\,\xi$ with $\xi\sim\mathcal{N}(0,I)$, where $\alpha$ is the perturbation magnitude relative to the score.
Performance remains stable up to $\alpha=0.1$ and degrades gradually beyond it (\cref{tab:appendix_correction}b).

\begin{table}[h]
\centering
\caption{\textbf{Terminal cost correction} (SANA 1.0). (a) Removing the correction. (b) Perturbing the estimated terminal score with relative magnitude $\alpha$.}
\label{tab:appendix_correction}
\begin{minipage}[t]{0.38\linewidth}
\centering
\textbf{(a)} Removal\\[0.3em]
\begin{tabular*}{\linewidth}{@{\extracolsep{\fill}}lcc@{}}
\toprule
& Ours & Reward only \\
\midrule
PickScore & \textbf{23.30} & 21.68 \\
\bottomrule
\end{tabular*}
\end{minipage}\hfill
\begin{minipage}[t]{0.58\linewidth}
\centering
\textbf{(b)} Perturbation\\[0.3em]
\begin{tabular*}{\linewidth}{@{\extracolsep{\fill}}lccccc@{}}
\toprule
$\alpha$ & 0 & 0.05 & 0.1 & 0.3 & 0.5 \\
\midrule
PickScore & 23.30 & 23.43 & 23.31 & 23.05 & 22.85 \\
\bottomrule
\end{tabular*}
\end{minipage}
\end{table}

\paragraph{Sampling NFE and LoRA rank.}
On SANA 1.0, we vary the number of ODE steps used to sample $X_1$ during training and the LoRA rank, with $C=0.51$ fixed.
As shown in~\cref{tab:appendix_nfe_rank}, EAM remains stable across all settings, and performance improves with both larger NFE and higher rank.
\begin{table}[H]
\centering
\caption{\textbf{Sampling NFE and LoRA rank} (SANA 1.0). Default settings are underlined.}
\label{tab:appendix_nfe_rank}
\begin{minipage}[t]{0.48\linewidth}
\centering
\textbf{(a)} Sampling NFE\\[0.3em]
\begin{tabular*}{\linewidth}{@{\extracolsep{\fill}}lccc@{}}
\toprule
NFE & 5 & \underline{10} & 20 \\
\midrule
PickScore & 23.15 & 23.30 & 23.41 \\
\bottomrule
\end{tabular*}
\end{minipage}\hfill
\begin{minipage}[t]{0.48\linewidth}
\centering
\textbf{(b)} LoRA rank\\[0.3em]
\begin{tabular*}{\linewidth}{@{\extracolsep{\fill}}lccc@{}}
\toprule
Rank & 16 & \underline{32} & 64 \\
\midrule
PickScore & 23.07 & 23.30 & 23.42 \\
\bottomrule
\end{tabular*}
\end{minipage}
\end{table}

\section{Additional Qualitative Examples}
\label{appendix:examples}
% \vspace{-2em}
% \vspace{-1em}
\begin{figure}[H]
    \centering
    \includegraphics[width=1.\linewidth]{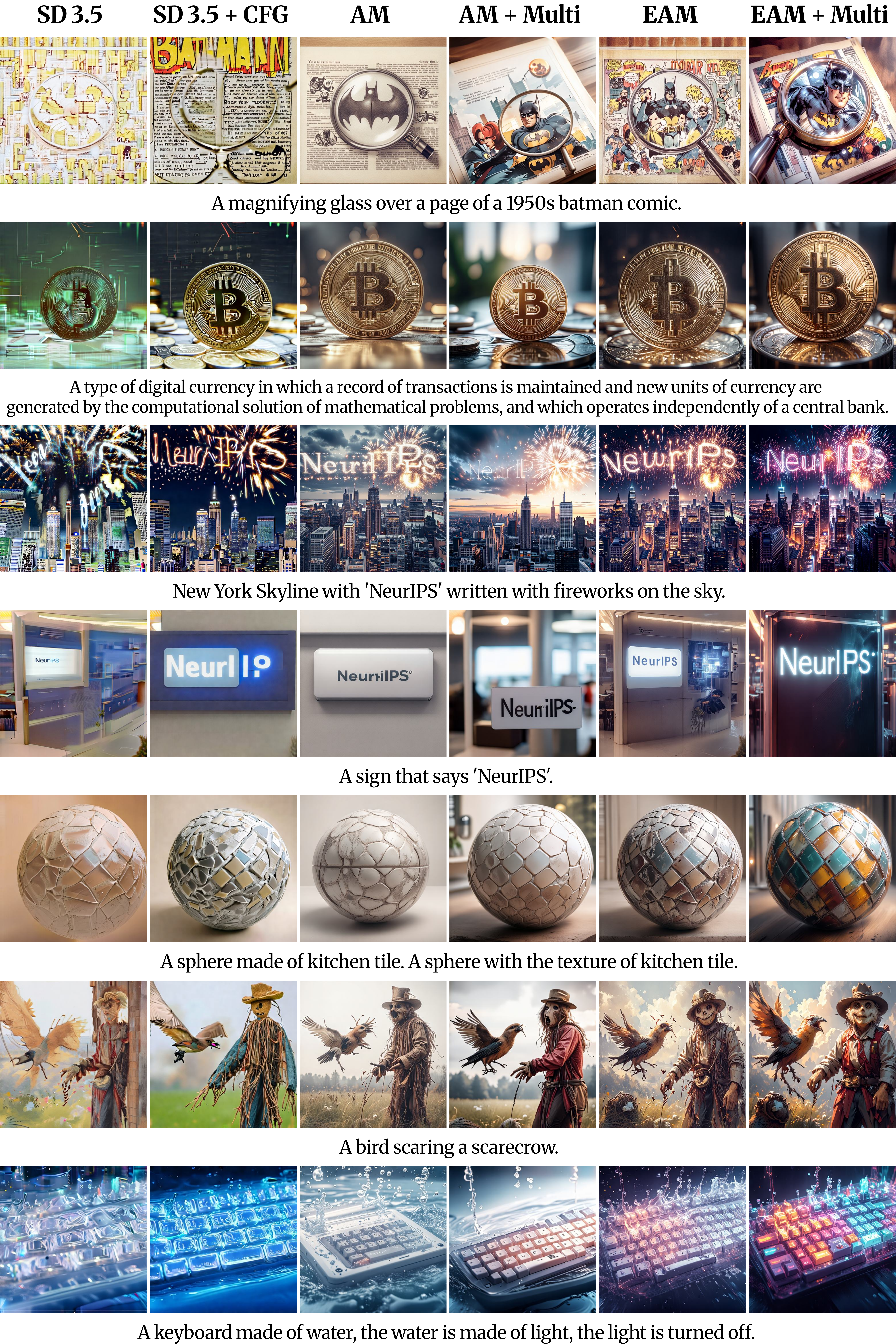}
    \vspace{-.4cm}
    \caption{
        \textbf{Additional qualitative results.} 
        % AM and EAM denote models fine-tuned using PickScore, while AM + Multi and EAM + Multi denote models fine-tuned using a combination of PickScore, HPSv2.1, and Aesthetics.
    }
    \vspace{-.4em}
\end{figure}

\begin{figure}[t]
    \centering
    \includegraphics[width=1.\linewidth]{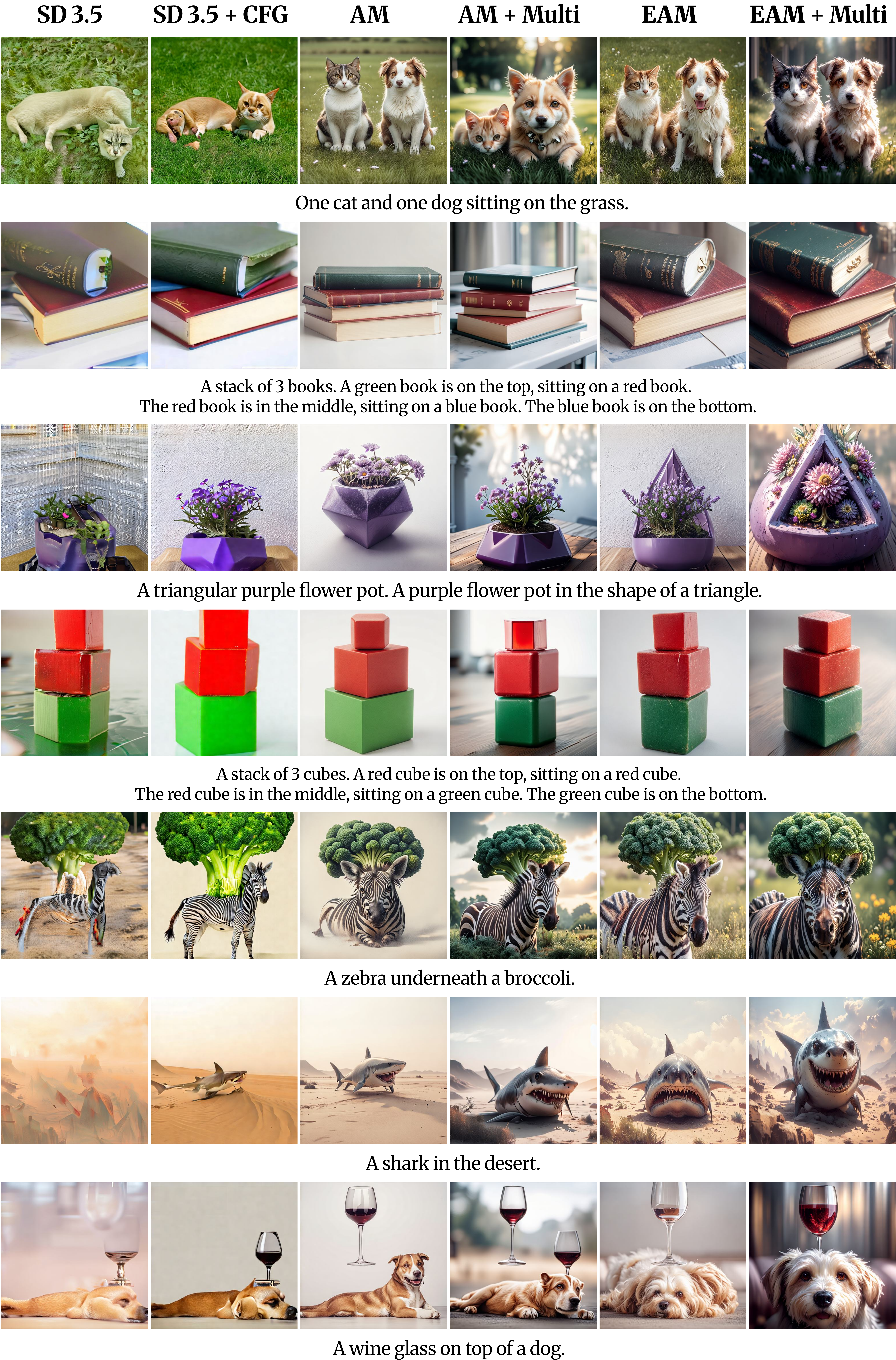}
    \vspace{-.4cm}
    \caption{
        \textbf{Additional qualitative results.} 
        % AM and EAM denote models fine-tuned using PickScore, while AM + Multi and EAM + Multi denote models fine-tuned using a combination of PickScore, HPSv2.1, and Aesthetics.
    }
    \vspace{-.4em}
\end{figure}

\begin{figure}[t]
    \centering
    \includegraphics[width=1.\linewidth]{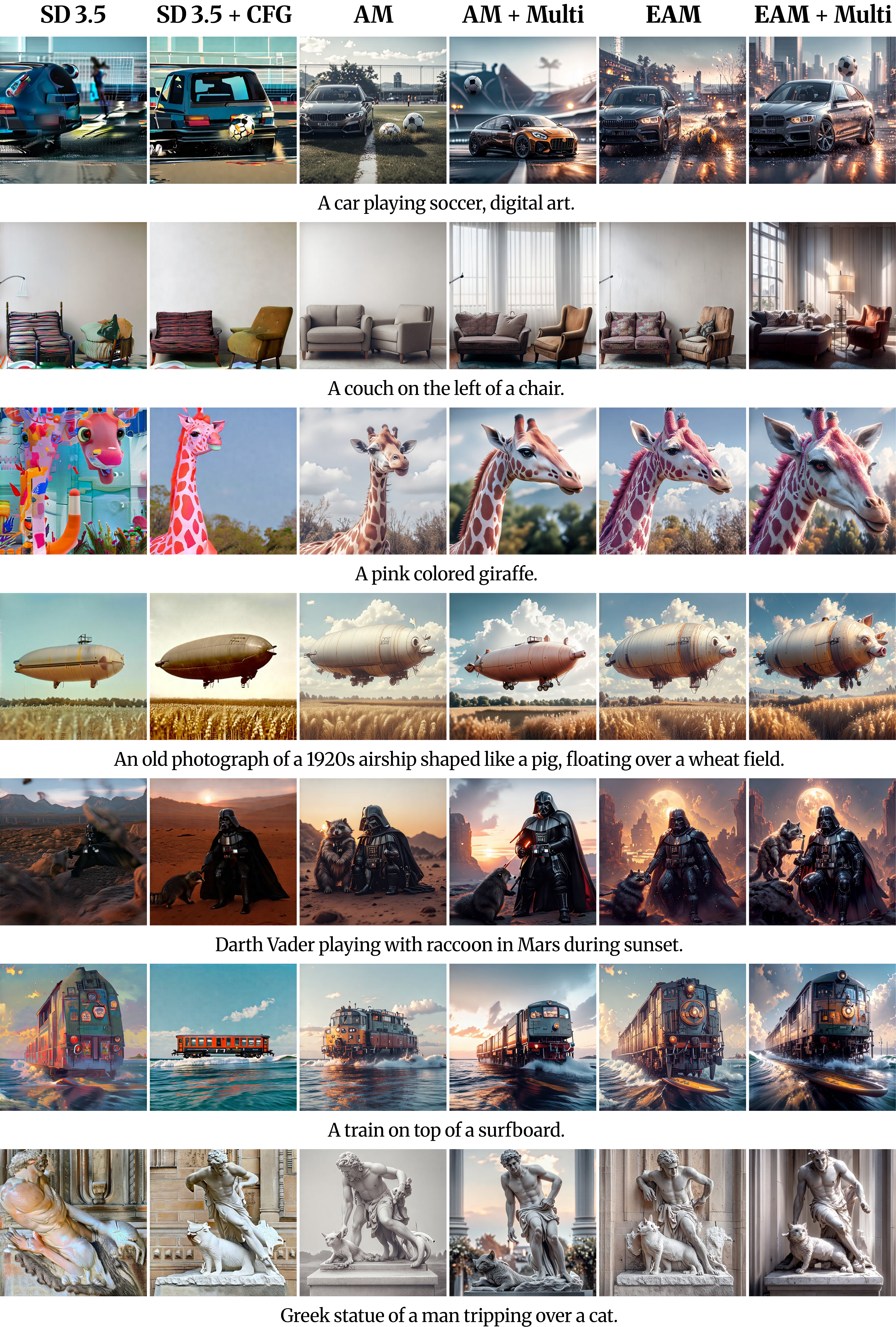}
    \vspace{-.4cm}
    \caption{
        \textbf{Additional qualitative results.} 
        % AM and EAM denote models fine-tuned using PickScore, while AM + Multi and EAM + Multi denote models fine-tuned using a combination of PickScore, HPSv2.1, and Aesthetics.
    }
    \vspace{-.4em}
\end{figure}

\end{document}